%% file: main.tex
\documentclass[11pt, letterpaper]{arxiv}

\usepackage[all]{hypcap}
\usepackage[comma,numbers,sort,compress]{natbib}
\usepackage[capitalize,noabbrev]{cleveref}
\usepackage{graphicx}
\usepackage{subfigure}
\usepackage{float}
\usepackage{wrapfig}
\usepackage{placeins}

\usepackage{booktabs}
\usepackage{colortbl}
\usepackage{multirow}
\usepackage{tabularx}
\newcommand{\tabledescript}{The \textcolor{gray}{[bracketed values]} represent a 95\% bootstrap confidence interval. The aggregate mean, median and interquartile mean (IQM) are computed over the }

\usepackage{amsmath}
\usepackage{amssymb}
\usepackage{mathtools}
\usepackage{amsthm}

\usepackage{algorithm}
\usepackage{algpseudocode}

\usepackage{microtype}
\usepackage{setspace}
\usepackage{xcolor}
\usepackage[most]{tcolorbox}

\usepackage{enumitem}

\usepackage{listings}
\usepackage{color}

\definecolor{sb_blue}{RGB}{65,105,225}
\definecolor{sb_orange}{RGB}{255,140,0}
\definecolor{gray}{RGB}{128,128,128}

\lstdefinestyle{mystyle}{
    language=Python,
    xleftmargin=5.0ex,
    basicstyle=\footnotesize\ttfamily\linespread{4},
    backgroundcolor=\color{gray!10},
    commentstyle=\color{gray},
    alsoletter={<>-0123456789},
    keywordstyle=\color{sb_blue},
    ndkeywords={nn, F, torch, partial},
    ndkeywordstyle=\color{sb_orange},
    emph={import, def, return, if, else},
    emphstyle=\bfseries\color{sb_blue},
    numberstyle=\footnotesize\ttfamily\color{gray},
    stringstyle=\color{sb_blue},
    breakatwhitespace=false,
    breaklines=true,
    keepspaces=true,
    numbers=left,
    numbersep=5pt,
    showspaces=false,
    showstringspaces=false,
    showtabs=false,
    tabsize=2
}
\definecolor{lightblue}{rgb}{0.22,0.45,0.70}
\newcommand{\Appendix}{\textcolor{mylinkcolor}{Appendix}}
\newcommand{\Figure}{\textcolor{mylinkcolor}{Figure}}
\newcommand{\Section}{\textcolor{mylinkcolor}{Section}}
\newcommand{\Table}{\textcolor{mylinkcolor}{Table}}

\title{Beyond Isolation: Unlocking Reinforcement Learning Component Synergy for Sample-Efficient Continuous Control}

\author[1]{Qi Zhao}
\author[2]{Guozheng Ma}
\author[2]{Yilun Kong}
\author[3]{Lu Li}
\author[2]{Haoyu Wang}
\author[4]{Zilin Wang}
\author[1]{Tiantian Zhang}
\author[1]{Yuxing Wang}
\author[1]{Jian Sha}
\author[1]{Yongzhe Chang}
\author[1]{Xueqian Wang}
\author[2]{Dacheng Tao}

\affil[1]{Tsinghua University}
\affil[2]{Nanyang Technological University}
\affil[3]{Mila - Quebec Artificial Intelligence Institute}
\affil[4]{University of Oxford}

\begin{abstract}
\input{Section/0_Abs}   
\end{abstract}

\begin{document}
\maketitle

\input{Section/1_Intro}          
\input{Section/2_Preliminary}    
\input{Section/3_Investigation}

\input{Section/5_Syn_Framework}
\input{Section/6_Experiments}
\input{Section/9_Conclusion}

\clearpage
\bibliography{ref}               

\newpage
\appendix
\onecolumn
\input{Section/Appendix}

\end{document}

%% file: Section/0_Abs.tex
Reinforcement learning systems are significantly more complex than other machine learning paradigms due to inherent properties, causing RL system design to jointly account for many tightly coupled factors.
Despite advances in individual algorithmic components, their functional interdependencies remain underexplored: \textbf{\textit{do they exhibit mutual synergy or counterproductive interference?}}
To bridge this gap, we conduct a systematic investigation and find that the efficacy of different components exhibits significant task-dependency, and naively stacking state-of-the-art techniques does not necessarily yield performance gains; instead, it often triggers emergent challenges, such as compounded non-stationarity. Building upon these findings, we distill a suite of actionable insights into the principled coordination of these components.
Guided by these insights, we propose \textbf{\texttt{ROSER}}, an RL framework that coordinates three critical dimensions: Model-based \textbf{\texttt{R}}epresentation, \textbf{\texttt{O}}ptimization \textbf{\texttt{S}}tability, and \textbf{\texttt{E}}xperience \textbf{\texttt{R}}eplay.
Across diverse continuous-control benchmarks, \textbf{\texttt{ROSER}} consistently outperforms vanilla baselines and achieves \textbf{~17.60\%} gains over naive stack.
Our findings underscore the necessity of a holistic perspective in RL system design and paves the way for developing sample-efficient agents.

%% file: Section/1_Intro.tex
\section{Introduction}
\label{sec_introduction}

\begin{figure}[t]
\centering
\begin{center}
\includegraphics[width=\linewidth]{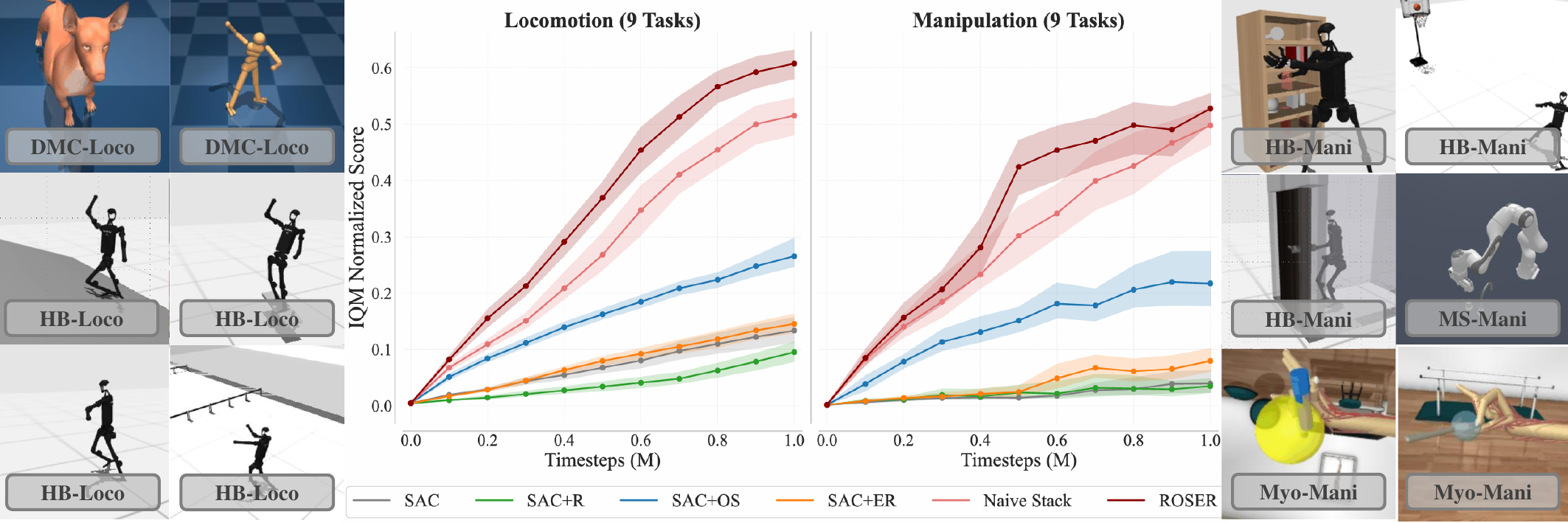}
\end{center}
\caption{\textbf{Performance Summary.} We compare \textbf{\texttt{ROSER}} with vanilla SAC, single-component enhancements (SAC+\textbf{\texttt{R}}, SAC+\textbf{\texttt{OS}}, SAC+\textbf{\texttt{ER}}), and a naive stack of all components (Naive Stack). \textbf{\texttt{ROSER}} demonstrates improved sample efficiency in both locomotion and manipulation tasks. Shaded regions represent the 95\% confidence interval. Experimental tasks span a diverse range of control scenarios, such as legged locomotion, dexterous object manipulation, articulated robotic arms.}
\label{sample_efficiency}
\end{figure}

Reinforcement learning (RL) aims to optimize an agent's decision-making strategy through environmental interaction. However, acquiring transitions often involves substantial time, computational resources, and, in many real-world applications, high physical or economic costs~\cite{kong2025qpo}. Consequently, improving sample efficiency has become a central challenge in RL research~\cite{yu2018towards, kong2025mastering}. Recent literature addresses this from many perspectives like: representation learning~\cite{fujimoto2023for, fujimoto2025towards} to enhance value and policy learning; optimization stability~\cite{nauman2024bigger, lee2025simba, lee2025hyperspherical} focus on improving the optimization dynamics of deep learning; prioritized sampling~\cite{schaul2015prioritized, fujimoto2020equivalence, lahire2021large} accelerate learning by assigning higher weights to more important transitions; and advanced exploration strategies~\cite{mahankali2024random, sukhija2025maxinforl} designed to collect more informative and diverse data. By refining these components, researchers aim to minimize the data required to reach optimal performance.

Despite significant advancements, most existing research focuses on improving sample efficiency through isolated components of the learning pipeline. 
It remains unclear whether different techniques for improving sample efficiency exhibit synergy, enhancing or counteracting each other. Furthermore, whether the joint integration of these techniques requires additional, specialized design considerations is a critical question that has yet to be explored.
Addressing these questions is of dual importance. From a research perspective, a deeper understanding of these interdependencies can provide systematic guidance for developing next-generation sample-efficiency techniques. From a practical standpoint, understanding these interactions is essential for building unified frameworks that truly maximize sample efficiency, enabling RL agents to master complex, challenging tasks with minimal environment interaction.

This work explores the joint enhancement paradigm and investigates the following core research questions: \textit{How do different sample-efficiency-enhancing techniques interact in reinforcement learning, and under what conditions can they be effectively combined to yield consistent and robust performance?} We build our investigation around three representative components: \textbf{Model-based Representation (\texttt{R})}, \textbf{Optimization Stability (\texttt{OS})}, and \textbf{Experience Replay (\texttt{ER})} mechanisms. These components are widely adopted in practice, though other components could be considered in future research.

Our primary finding underscores the necessity of investigating sample efficiency within integrated, high-performance contexts rather than in isolation on top of basic algorithms (e.g., vanilla SAC). Specifically, we observe that the efficacy of existing enhancement techniques exhibits significant task-dependency; a mechanism that provides substantial gains in some tasks may cause performance degradation in others. Furthermore, our investigation reveals that simply stacking individual techniques does not necessarily yield an improvement in sample efficiency. This phenomenon suggests that the integration of multiple enhancement strategies introduces emergent challenges, such as mutual interference and compounded non-stationarity, that are absent when modules are studied in isolation. Our investigation yields three design insights for harmonizing multiple enhancements. 
Guided on these insights, we present \textbf{\texttt{ROSER}}, a reinforcement learning framework designed for synergistic enhancement. \textbf{\texttt{ROSER}} harmoniously integrates representative techniques from three critical components: \textbf{\texttt{R}}, \textbf{\texttt{OS}}, and \textbf{\texttt{ER}}. As shown in \Figure~\ref{sample_efficiency}, \textbf{\texttt{ROSER}} achieves a notable gain in sample efficiency, outperforming methods that rely on isolated component enhancements and simply stacking individual techniques by a significant margin.

\vspace{-0.5\baselineskip}
\contribution{
Contributions of this paper can be summarized as:
\vspace{-0.2\baselineskip}
\begin{enumerate}[leftmargin=*, itemsep=0em]
\item \textbf{Interaction Investigation:} We conduct an analysis of various techniques, revealing that their individual benefits do not scale linearly and can even exhibit counter-productive interference in integrated systems, demonstrating that the coordination problem is real and non-trivial.
\item \textbf{Design Principle:} Informed by our analysis, we distill three design principles to harmonize algorithmic interactions. Based on these, we propose \textbf{\texttt{ROSER}}, a reinforcement learning framework that achieves superior sample efficiency and cross-environment versatility.
\item \textbf{Empirical Validation:} Extensive experiments demonstrate the superior performance of \textbf{\texttt{ROSER}}, validating our distilled design principles and underscoring that the principled coordination of algorithmic components is essential for achieving peak sample efficiency.
\end{enumerate}
}
\vspace{-0.75\baselineskip}

%% file: Section/2_Preliminary.tex
\section{Background}
\label{sec_preliminary}

This section provides the background and formal notation for our study. We review previous research that takes a system-level perspective to reinforcement learning frameworks, and present a focused overview of three representative components, \textbf{\texttt{R}}, \textbf{\texttt{OS}}, and \textbf{\texttt{ER}}.

\subsection{Systematic Perspectives for RL}

The systematic integration of RL modules was pioneered by \textbf{Rainbow}~\cite{hessel2018rainbow}, which combined several extensions to achieve state-of-the-art performance. \textbf{Revisiting Rainbow}~\cite{ceron2021revisiting} expanded on this work by advocating for inclusive evaluations across diverse regimes. More recently, \textbf{Beyond The Rainbow (BTR)}~\cite{clark2025beyond} further advanced this paradigm by integrating six modern algorithmic and architectural enhancements, establishing a new state-of-the-art with desktop-level computational efficiency. While these studies established the potential of component aggregation, they primarily focused on value-based methods (DQN) in discrete action spaces and treated modules as additive "plug-ins" without deeply exploring the underlying intervention between them.

Our study builds upon this systematic lineage but shifts the focus toward \textbf{Component Synergy}, specifically within actor-critic frameworks (e.g., SAC) in continuous action spaces. We examine the interaction among some components to determine if they show synergy or interference. Beyond mere identification, we provide a diagnostic analysis of the underlying mechanisms that drive negative interactions, offering insights into component relationships.

\subsection{Model-based Representation (\texttt{R}) for RL}
Representation learning aims to extract useful features from high-dimensional sensory inputs (e.g., images, raw observations) that are relevant for decision-making~\cite{bengio2014representation}. A good representation should retain essential information for decision-making while discarding irrelevant details. To achieve this, modern RL frameworks typically employ auxiliary tasks to provide additional supervision for the encoder network. 
By leveraging these self-supervised signals, such as reconstructing observations \cite{ha2018world}, predicting future latent states \cite{Hafner2020Dream}, or using contrastive estimation \cite{srinivas2020curl}, the agent can "squeeze" significantly more information out of every transition collected from the environment, which substantially boosts sample efficiency.

\textbf{MR.Q}~\cite{fujimoto2025towards} learns a state embedding $z_s$ and a state-action embedding $z_{sa}$ through end-to-end training of an encoder. The encoder loss $L_{enc}$ is formulated as:
\begin{equation}
L_{enc} = \lambda_1\cdot L_{rew}+\lambda_2\cdot L_{dyn}+\lambda_3\cdot L_{term}
\end{equation}
where $L_{rew}$, $L_{dyn}$, $L_{term}$ are the losses corresponding to the reward, the dynamics, and the terminal state prediction tasks, respectively. The terms $\lambda_1$, $\lambda_2$ and $\lambda_3$ are weighting factors that balance the contribution of each auxiliary task.
Furthermore, this representation learning is decoupled from downstream RL training and updated periodically to ensure a stable feature space. We adopt this MR.Q-style learning process due to its solid theoretical grounding and its model-based formulation, which produces a more structured and optimization-friendly latent space. For detailed implementation, refer to \Appendix~\ref{appendix_encoder}.

\subsection{Optimization Stability (\texttt{OS}) for RL}

In recent years, an increasing body of work has focused on the network pathologies encountered in deep reinforcement learning (DRL). Among the most severe of these pathologies are capacity collapse and plasticity loss~\cite{ma2025rethinkingroledynamicsparse}. Plasticity loss~\cite{klein2024plasticitylossdeepreinforcement} refers to a phenomenon in which a neural network, trained on a sequence of non-stationary distributions, progressively loses its ability to adapt to new data. This issue is particularly problematic in the context of RL, where the agent's evolving policy continuously shifts the data distribution stored in the replay buffer. If the network becomes excessively specialized to earlier experiences (referred to as "primacy bias"~\cite{nikishin2022primacy}), it may fail to incorporate more optimal behaviors that emerge later in the training process. Recent research suggests that certain network architectures, such as layer normalization and residual connections, can be particularly helpful in mitigating these pathologies~\cite{nauman2024bigger}. In this regard, SimBa~\cite{lee2025simba} has conducted extensive analysis and experiments, proposing a network architecture that is both simple and effective.

\textbf{SimBa}~\cite{lee2025simba} addresses optimization stability by amplifying simplicity bias into the architecture of deep RL models. By constraining the network's architecture to favor simpler representations, SimBa reduces the risk of overfitting to noise in the training data. Given the empirically proven effectiveness of SimBa across various experimental settings, we adopt its architecture as the representative technique for \textbf{\texttt{OS}} in our study. Detailed architectural specifications are provided in \Appendix~\ref{appendix_simba_block}, while the hyperparameter configurations are listed in \Table~\ref{appendix:OS_hparams}. 

\subsection{Experience Replay (\texttt{ER}) for RL}

Experience replay is a cornerstone of off-policy RL, allowing an agent to store and reuse past experiences. The idea is to sample transitions from a buffer \(\mathcal{D}\) to break the temporal correlation of consecutive states and actions, which reduces the variance in updates and improves the stability of training~\cite{fedus2020revisitingfundamentalsexperiencereplay}. One challenge with experience replay is ensuring that the sampling process is efficient. Uniform sampling of experiences may not fully exploit the most informative transitions. Instead, prioritizing samples based on their potential can help improve data efficiency. However, TD-error-based prioritization~\cite{schaul2015prioritized, fujimoto2020equivalence, lahire2021large} often suffers from "wasted" gradients: large TD errors may arise from stochasticity or irreducible noise, causing the agent to repeatedly sample transitions that are effectively unlearnable~\cite{sujit2022prioritizing}. Efficient \textbf{\texttt{ER}} should therefore distinguish genuinely informative samples from the replay buffer.

\textbf{ReLo}~\cite{sujit2022prioritizing} introduces the concept of \textit{Reducible Loss (ReLo)}, which prioritizes data points that most reduce the model's generalization loss. The key idea is to train the model on a subset of the data and compare its performance with a \textit{hold-out model} trained without the current data point. The \textit{Reducible Loss} $L_r$ for a data point $x_i$ is defined as:
\begin{equation}
L_r = \text{Loss}(\hat{y} | x, \theta) - \text{Loss}(\hat{y} | x, \theta_h)
\end{equation}
where \(\hat{y}\) is the predicted output, \(\theta\) are the parameters of the main model, and \(\theta_h\) are the parameters of the hold-out model. ReLo down-weights unlearnable transitions, thereby enhancing the sample efficiency.

Following a comparative analysis of several experience replay techniques, including PER~\cite{schaul2015prioritized}, LAP~\cite{fujimoto2020equivalence}, LaBER~\cite{lahire2021large}, and ReLo~\cite{sujit2022prioritizing} (see \Appendix~\ref{appendix_analysis}), we adopt ReLo as the primary representative technique for \textbf{\texttt{ER}} in our subsequent experiments and analysis.

%% file: Section/3_Investigation.tex
\section{Investigation}
\label{sec_investigation}

In this section, we conduct a systematic investigation on continuous-control tasks to examine the interactions among various sample-efficiency-enhancing techniques when integrated into a framework. Our analysis begins by investigating the performance enhancements brought by \textbf{\texttt{OS}} to other components. Subsequently, we transition to the integration of \textbf{\texttt{R}}, focusing on how to unlock their full potential. Finally, we present the dilemma of incorporating \textbf{\texttt{ER}} and propose targeted solutions to mitigate this issue.

\textbf{Experiment Setup}~To enable a comprehensive and reliable empirical study and to strengthen the robustness of our conclusions, we evaluate on a diverse set of continuous-control tasks drawn from four widely adopted and authoritative benchmarks: \textit{\textbf{DeepMind Control suite (DMC)}}~\cite{tassa2018dmc}, \textit{\textbf{HumanoidBench (HB)}}~\cite{sferrazza2024humanoidbench}, \textit{\textbf{Myosuite (Myo)}}~\cite{MyoSuite2022}, and \textit{\textbf{ManiSkill2 (MS)}}~\cite{gu2023maniskill2}. Rather than exhaustively covering all tasks in these environments, we focuse on tasks that remain challenging. Specifically, we consider a total of 18 tasks, consisting of 9 locomotion tasks and 9 manipulation tasks, covering a broad range of control scenarios, as visually depicted in \Figure~\ref{sample_efficiency}. Additional environment-specific details and task configurations are provided in \Appendix~\ref{appendix_environments}. We conduct every experiment across 8 random seeds. 

We adopt Soft Actor-Critic (SAC) \cite{haarnoja2018soft} as the base algorithm, upon which various sample-efficiency-enhancing techniques are integrated and evaluated. Notably, we utilize a fixed set of hyperparameters across all experiments in this paper without any specific tuning. The complete hyperparameter settings are detailed in \Appendix~\ref{appendix_hyperparameters}. Following the recommendations of \citet{agarwal2021deep}, we report the Interquartile Mean (IQM) with 95\% stratified bootstrap confidence intervals to provide a robust assessment of performance across heterogeneous environments. The IQM is computed based on the maximum return or maximum success rate achieved during the training process.

\subsection{Optimization Stability Enables Synergy Across Sample-Efficiency Modules}

\begin{wrapfigure}{r}{0.5\textwidth}
\vspace{-1.0\baselineskip}
\centering
\includegraphics[width=0.48\textwidth]{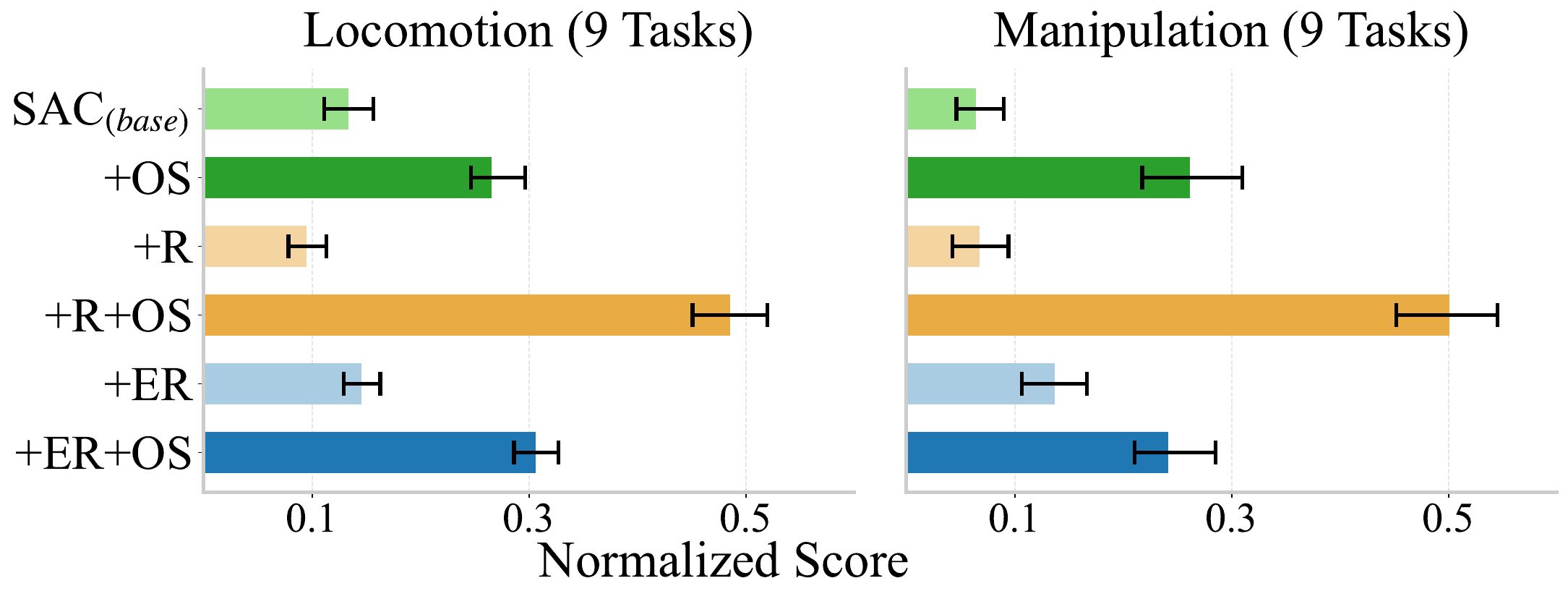}
\vspace{-0.5\baselineskip}
\caption{\textbf{Performance Gains of Optimization Stability.} The results demonstrate that the Simba backbone consistently enhances performance.}
\vspace{-1.5\baselineskip}
\label{exp_os_effect}
\end{wrapfigure}

To investigate the role of \textbf{\texttt{OS}} in coordination with other sample-efficiency-enhancing techniques, we integrate the Simba network architecture with various modules to examine whether a stable backbone is a prerequisite for effective component synergy. Specifically, we focus on three comparisons that isolate the incremental effect of Simba under different settings: (i) SAC vs. SAC + \textbf{\texttt{OS}}; (ii) SAC + \textbf{\texttt{R}} vs. SAC + \textbf{\texttt{R}} + \textbf{\texttt{OS}}; and (iii) SAC + \textbf{\texttt{ER}} vs. SAC + \textbf{\texttt{ER}} + \textbf{\texttt{OS}}.

As illustrated in \Figure~\ref{exp_os_effect}, the Simba backbone provides a consistent performance boost across all evaluated configurations.
In all cases, we observe a domain-agnostic synergistic amplification effect: when integrated with other sample-efficiency-enhancing techniques, Simba consistently yields a positive synergy.
The results in case (ii) are noteworthy, as they reveal that introducing \textbf{\texttt{R}} in isolation yields a negative impact, particularly on locomotion tasks. However, when integrated with \textbf{\texttt{OS}}, performance improves drastically, surpassing the individual capabilities of either component by a significant margin. This not only demonstrates the generalizable benefits of \textbf{\texttt{OS}} across diverse RL settings, but further highlights how proper component integration can trigger super-additive, "1+1>2" synergistic gains.

\vspace{-0.25\baselineskip}
\Takeaway{Our results demonstrate that integrating Simba with various modules consistently yields performance gains. This underlines a broader architectural insight: a stable optimization network backbone is essential for fostering effective synergy among components.}
\vspace{-0.25\baselineskip}

\subsection{Stabilizing Information Flow to Strengthen Component Synergy}

\begin{figure*}[ht]
\centering
\begin{minipage}{0.48\textwidth}
    \centering
    \includegraphics[width=\linewidth]{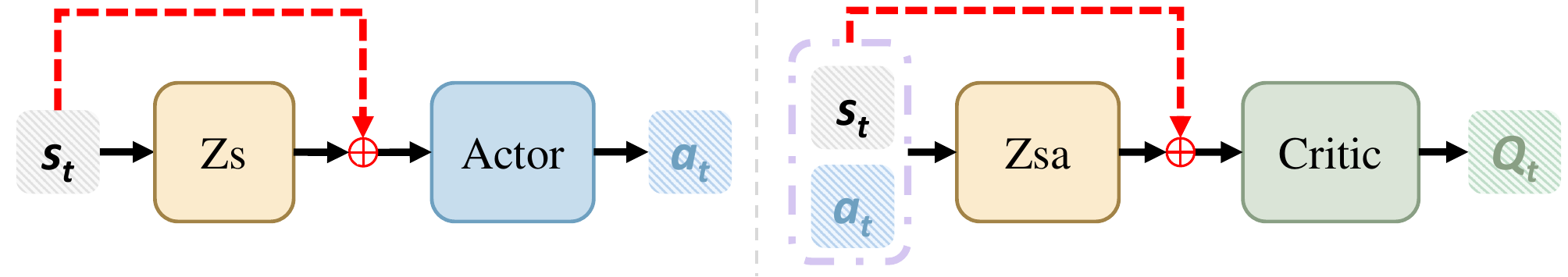}
    \vspace{-0.5\baselineskip}
    \caption{\textbf{Information Flow of \textbf{\texttt{R}$\star$}.} Our scheme concatenates raw state $s_t$ and action $a_t$ with learned features via \textbf{residual-like connections}.}
    \label{pre_representation}
\end{minipage}
\hfill
\begin{minipage}{0.48\textwidth}
    \centering
    \includegraphics[width=\linewidth]{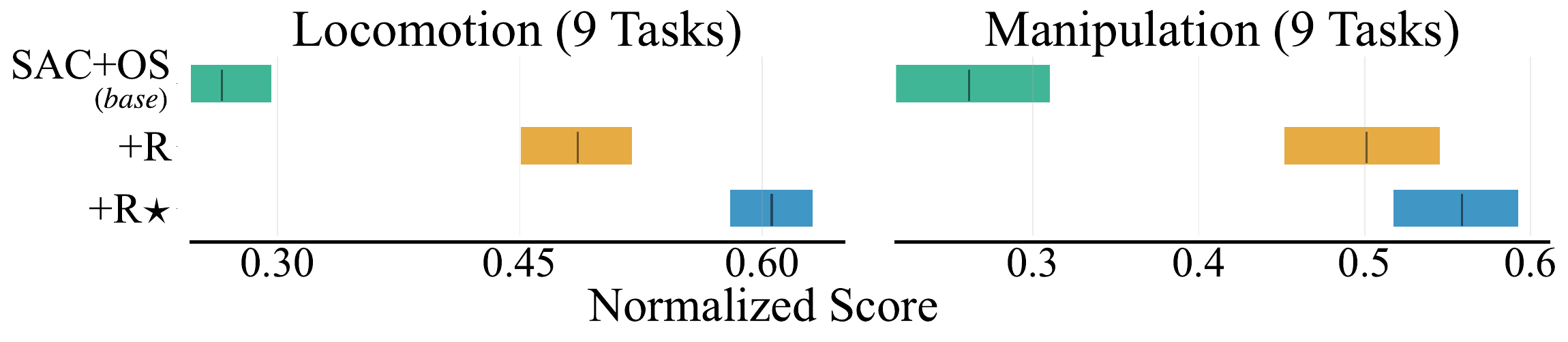}
    \vspace{-1.2\baselineskip}
    \caption{\textbf{Comparison of Representation Schemes.} We compare the baseline, naive integration (+\textbf{\texttt{R}}), and stabilized variant (+\textbf{\texttt{R}$\star$}).}
    \label{exp_ros_effect}
\end{minipage}
\vspace{-0.5\baselineskip}
\end{figure*}

Previous results demonstrate that the naive integration of Simba with model-based representation already yields notable performance gains. We argue that this empirical success can be partially attributed to the inherent advantages of residual connections, where shortcut connections fundamentally stabilize \textit{micro-level} information flow within neural architectures. Motivated by this insight, we extend this philosophy to the \textit{macro-level} and introduce a \textbf{Sta}ble Model-based \textbf{R}epresentation, denoted as \textbf{$\texttt{R}\star$} (pronounced as 'R-\textbf{StaR}'). The core design involves a residual-style feature fusion (\Figure~\ref{pre_representation}): learned latent embeddings are concatenated with original features before entering downstream policy and value networks. This architectural "information bypass" ensures that the actor and critic maintain access to stable signals.

We empirically evaluate the necessity of this design by comparing SAC + \textbf{\texttt{OS}} + \textbf{\texttt{R}} against its stabilized counterpart, SAC + \textbf{\texttt{OS}} + \textbf{$\texttt{R}\star$}. As shown in \Figure~\ref{exp_ros_effect}, the \textbf{$\texttt{R}\star$} variant significantly outperforms the naive combination, particularly in locomotion scenarios. Specifically, it achieves a performance boost of 24.72\% in locomotion tasks and 11.51\% in manipulation tasks compared to the naive combination components. These results underscore that while Simba provides a powerful backbone, the robustness of the information flow is crucial for maximizing the synergistic effects of model-based representation.

\Takeaway{\textbf{$\texttt{R}\star$} consistently pushing the performance boundaries of SAC+\textbf{\texttt{OS}}+\textbf{\texttt{R}} suggests that the stable information flow via residual-style bypasses is crucial for effective inter-modular synergy.}

\subsection{A Dilemma of Experience Replay in Coordinated Integration}

While non-uniform experience replay mechanisms are well-established for boosting sample efficiency in isolated vanilla settings, it remains an open and critical question how these sampling priorities behave when integrated into a more complex system. To this end, we introduce prioritized replay denoted as \textbf{\texttt{ER}}(\textbf{P}) into SAC + \textbf{\texttt{OS}} + \textbf{$\texttt{R}\star$}, yielding a variant referred to as SAC + \textbf{\texttt{OS}} + \textbf{$\texttt{R}\star$} + \textbf{\texttt{ER}}(\textbf{P}).

\begin{wrapfigure}{r}{0.5\textwidth}
\vspace{-0.8\baselineskip}
\centering
\includegraphics[width=\linewidth]{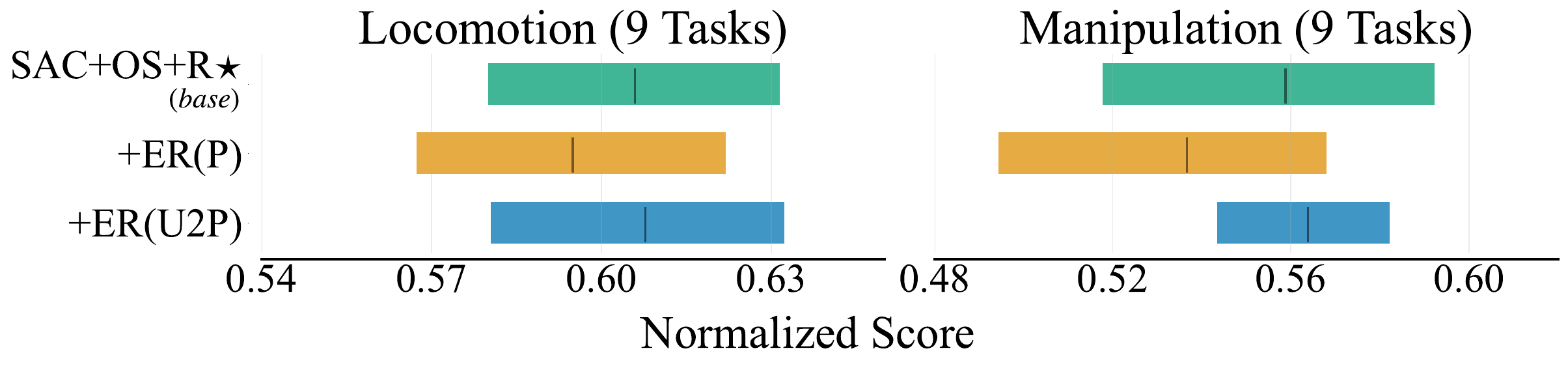}
\caption{\textbf{Comparison of Experience Replay Schemes.} We compare SAC+\textbf{\texttt{OS}}+\textbf{$\texttt{R}\star$} integrated with different experience replay variants, including \textbf{Prioritized replay} (\textbf{\texttt{ER}}(\textbf{P})) and \textbf{Uniform-to-Prioritized replay} (\textbf{\texttt{ER}}(\textbf{U2P})).}
\vspace{-3.0\baselineskip}
\label{exp_roser_effect}
\end{wrapfigure}

Contrary to expectations, \Figure~\ref{exp_roser_effect} shows that this configuration underperforms SAC + \textbf{\texttt{OS}} + \textbf{$\texttt{R}\star$}. This degradation indicates a counteractive effect between prioritized experience replay and the previously integrated components. These results highlight an important insight: \textit{naively stacking multiple sample-efficiency-enhancing techniques does not guarantee additive improvements}. We conjecture that, during the early stage of training, when representation learning and value estimation are still evolving, priority signals may become less reliable and introduce undesirable sampling biases.

Based on this consideration, we propose \textbf{Uniform-to-Prioritized replay (U2P)}. Instead of prioritized sampling throughout training, \textbf{U2P} schedules the exponent $\alpha$, defined as:
\begin{equation}
\alpha_t \!= \!\!\begin{cases}
0.0 & \!\!\!\!\!t\!<\!t_{\text{start}} \\
\alpha_F\!\left(\!1\!-\!\left(1\!-\!\tfrac{t-t_{\text{start}}}{t_{\text{end}}-t_{\text{start}}}\right)^{\!2}\right) & \!\!\!\!\!t_{\text{start}}\! \leq t \!\leq t_{\text{end}} \\
\alpha_F & \!\!\!\!\!t\!>\!t_{\text{end}}
\end{cases}
\end{equation}
Here, $\alpha$ controls the degree of non-uniform sampling: $P(i)=\frac{p_i^\alpha}{\sum_k p_k^\alpha}$. Detailed hyperparameters are provided in \Table~\ref{appendix:ER_hparams}. This design gradually increases the influence of prioritized replay, allowing the replay distribution to adapt progressively alongside the evolving learning process. As illustrated in \Figure~\ref{exp_roser_effect}, incorporating \textbf{U2P} successfully mitigates the conflict between \textbf{\texttt{ER}} with \textbf{$\texttt{R}\star$} and \textbf{\texttt{OS}}.

\Takeaway{Naively stacking \textbf{\texttt{ER}} on top of \textbf{$\texttt{R}\star$} and \textbf{\texttt{OS}} can be counterproductive. This highlights the importance of coordinating replay strategies with other components. A scheduled prioritization strategy, \textbf{U2P}, improves compatibility and enables more effective component integration.}

\textbf{Summary of Investigation}~Taking optimization stability, model-based representation, and experience replay as examples, this investigation presents the intricate interplay that emerges when these DRL components are jointly integrated. Crucially, we reveal that enhancements demonstrating remarkable individual success on vanilla SAC exhibit divergent behaviors when integrated. Specifically, we show that optimization stability is not merely beneficial in isolation, but serves as a foundational enabler that allows other components to synergize effectively. Yet micro-level network stability alone is insufficient; scaling this principle to a macro-level architecture by ensuring a reliable information flow for representation learning further catalyzes algorithmic gains. Conversely, advanced experience replay, despite its well-known efficacy in vanilla settings, presents conflicts when naively superimposed onto a complex configuration. Together, these results highlight that achieving strong sample efficiency is a systems problem: gains emerge from coherent co-design.

%% file: Section/5_Syn_Framework.tex
\section{Synergistic Framework}
\label{sec_syn_framework}

Building on the investigation, we distill a set of general principles aimed at effectively coordinating diverse sample-efficiency-enhancing techniques. We then consolidate these principles into a framework, termed \textbf{\texttt{ROSER}} (Model-based \textbf{\texttt{R}}epresentation, \textbf{\texttt{O}}ptimization \textbf{\texttt{S}}tability and \textbf{\texttt{E}}xperience \textbf{\texttt{R}}eplay). 
Through principled coordination, \textbf{\texttt{ROSER}} promotes positive effectiveness among these components, enabling them to amplify performance and robustness in challenging continuous-control tasks.

\subsection{Design Principles for Synergistic Coordination} 

This subsection introduces the design principles.

\textbf{\textcolor{mydarkgreen}{$\bullet$~Principle 1:} Optimization Stability Serves as the Foundational Backbone}~Our investigation indicates that optimization stability plays a foundational role in integrated systems. A stable optimization backbone establishes favorable learning dynamics that allow other components to express their potential. We therefore treat optimization stability as the structural backbone of the framework.

\textbf{\textcolor{mydarkgreen}{$\bullet$~Principle 2:} Synergy Requires Stable Information Flow} Residual connections are well-established as a stabilizing mechanism within neural networks. We demonstrate that this philosophy extends beyond the intra-network level, showing that explicit bypass connections can stabilize information flow and yield substantial gains in component synergy. We believe this principle is not confined to our specific design, and encourage the community to further explore residual-style bypasses as a general design guideline.

\textbf{\textcolor{mydarkgreen}{$\bullet$~Principle 3:} Synergy Favors Robust Coordination over Aggressive Individual Performance.} In integrated settings, overly aggressive prioritized experience replay can disrupt coordination among components and degrade performance. In contrast, more conservative and progressive designs tend to preserve compatibility. This highlights a key principle for synergistic integration: prioritize robustness of coordination over maximizing the standalone strength of any individual component.

\subsection{Instantiating the \texttt{ROSER} Framework}

Guided by the design principles, we instantiate \textbf{\texttt{ROSER}}. First, building on first design principle, we integrate the Simba network architecture across all functional modules of the agent, including encoder, actor, and critic (\Appendix~\ref{appendix_architecture}). Second, to fully harness the potential of representation learning within the integrated system, we instantiate the residual-style information bypass philosophy at the macro-level via \textbf{$\texttt{R}\star$}, ensuring that downstream policy and value networks retain direct access to stable input signals. Finally, when incorporating experience replay, we employ the \textbf{U2P} replay strategy to delay prioritization until learning signals become sufficiently reliable, ensuring that experience replay acts as a harmonizing force rather than a source of instability.

Our synergistic framework is algorithm-agnostic and can be seamlessly integrated into a wide range of off-policy continuous-control algorithms, DDPG and SAC. In this work, we primarily instantiate \textbf{\texttt{ROSER}} on top of SAC, while additional results on DDPG are provided in \Appendix~\ref{appendix_ddpg}.

%% file: Section/6_Experiments.tex
\section{Experiments}
\label{sec_experiments}

This section evaluates the effectiveness and applicability of the proposed \textbf{\texttt{ROSER}} framework on a diverse set of continuous-control tasks. Rather than establishing \textbf{\texttt{ROSER}} as a state-of-the-art algorithm competing against the strongest available baselines, our core objective is to demonstrate that principled coordination of sample-efficiency-enhancing components can yield substantially improved sample efficiency. To this end, we first benchmark \textbf{\texttt{ROSER}} against vanilla SAC, controlled partial variants, and naive stacking, assessing its sample efficiency and robustness across diverse environments. We then conduct a targeted analysis to examine the source of performance gains from \textbf{$\texttt{R}\star$} and \textbf{U2P}.

\subsection{Evaluation of \texttt{ROSER}}
\textbf{Experimental Setup}
The task configurations and environments used here are consistent with those employed in \Section~\ref{sec_investigation}.
We compare \textbf{\texttt{ROSER}} against the vanilla SAC algorithm and several partially enhanced variants that incorporate only a subset of its components:
(i) SAC with model-based representation (SAC+\textbf{\texttt{R}}),
(ii) SAC with optimization stability (SAC+\textbf{\texttt{OS}}),
(iii) SAC with enhanced experience replay (SAC+\textbf{\texttt{ER}}),
and (iv) Naive Stack framework, which naively stacks all above components. We report results across 8 stochastic seeds for each configuration.

\begin{wrapfigure}{r}{0.5\textwidth}
\centering
\vspace{-0.6\baselineskip}
\includegraphics[width=\linewidth]{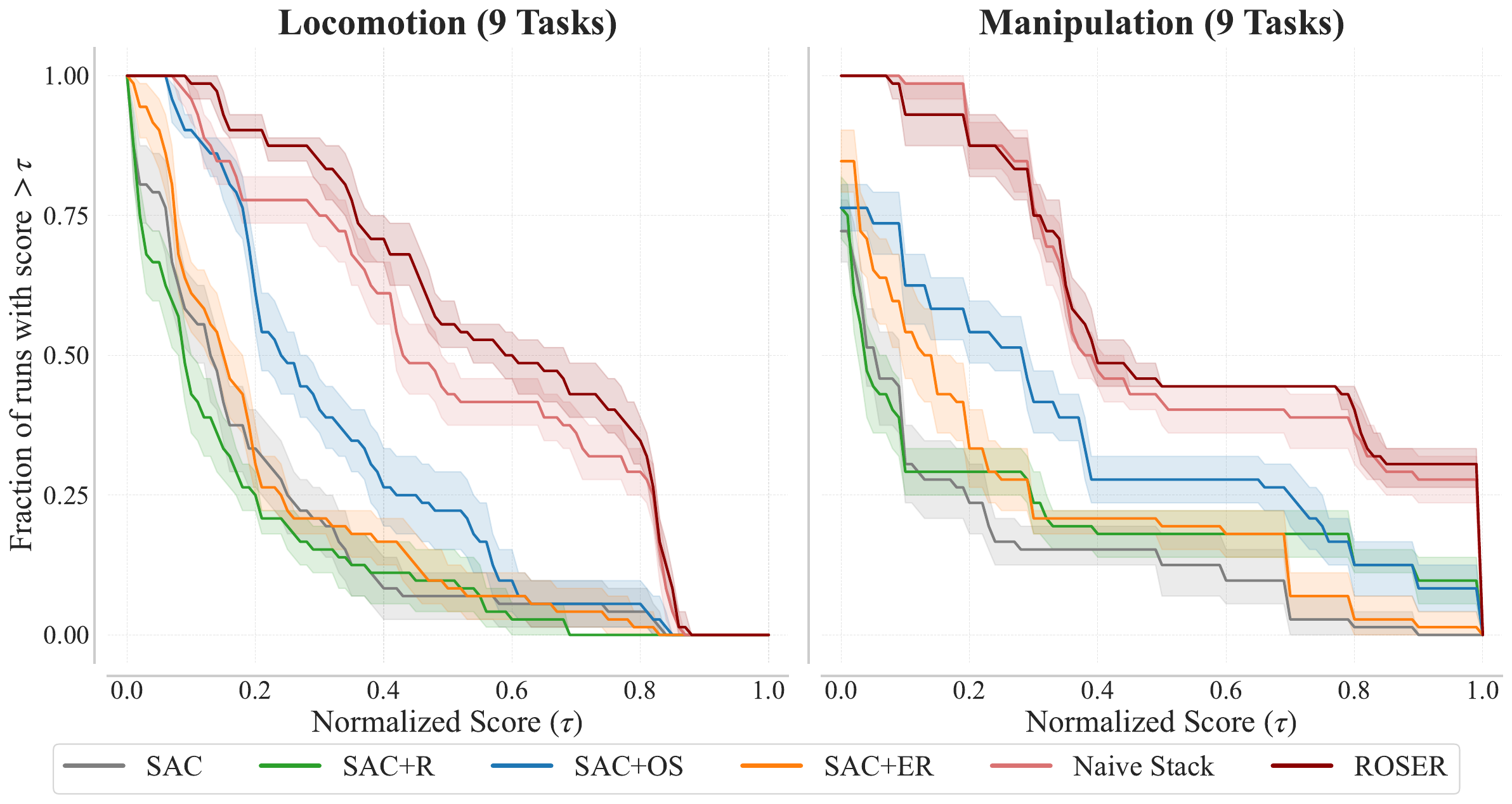}
\caption{\textbf{Comparison of Efficiency and Reliability.} Fraction of runs achieving normalized score $> \tau$; 95\% CI. Curves closer to top-right indicate superior performance and reliability.}
\vspace{-1\baselineskip}
\label{exp_main_performance_profiles}
\end{wrapfigure}

\begin{figure*}[t]
\centering
\begin{center}
\includegraphics[width=\linewidth]{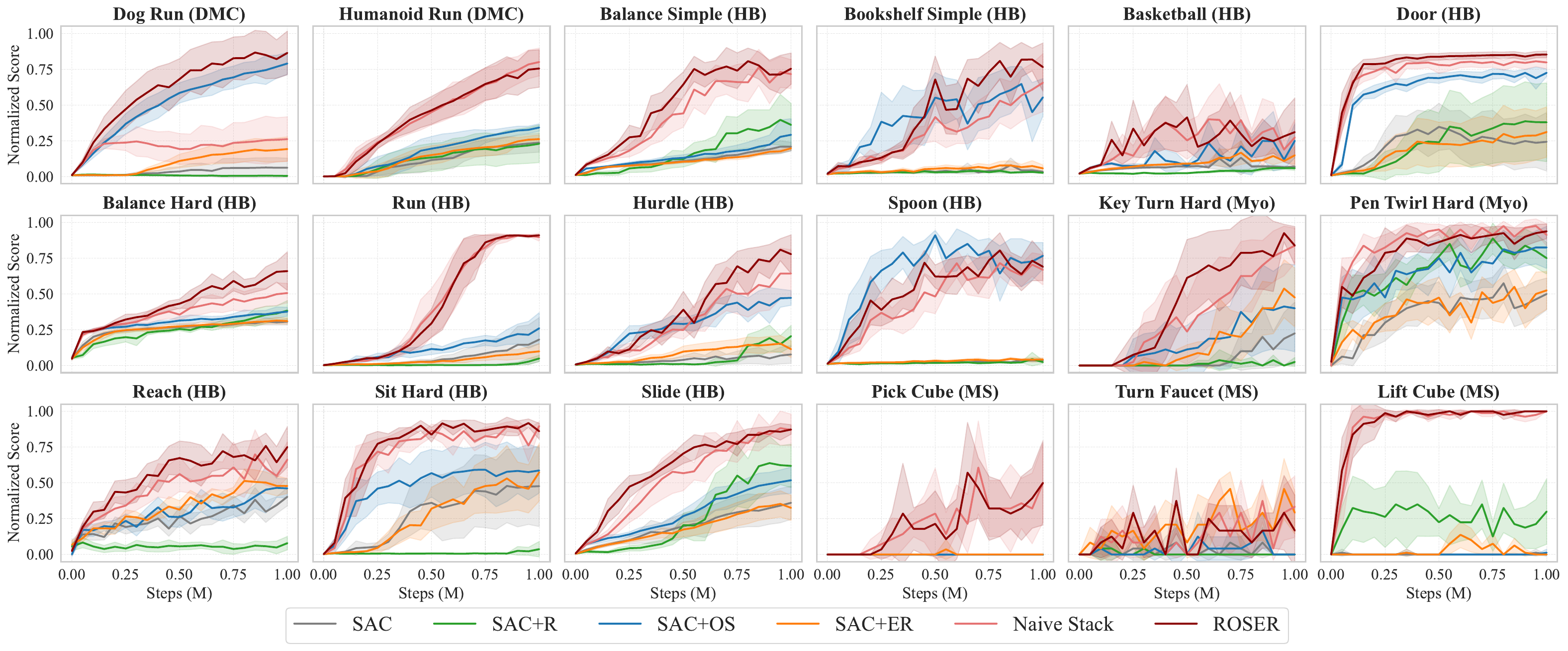}
\end{center}    
\caption{\textbf{Comparison of Performance Across Different Tasks.} We plot the learning curve of \textbf{\texttt{ROSER}} and baseline algorithms, categorized into locomotion (left 3$\times$3 panels) and manipulation (right 3$\times$3 panels). To better compare the performance differences, we normalize the scores by rounding up the highest score in each task to the nearest hundred, resulting in the \textit{Normalized Score}. Results without normalization can be found in \Appendix~\ref{appendix_results}. Shaded regions represent the 95\% confidence interval.}
\label{exp_main_learning_curve}
\end{figure*}

To rigorously evaluate the proposed algorithms, we adopt the statistical framework recommended by \citet{agarwal2021deep}. We first aggregate results across all tasks using performance profiles (\Figure~\ref{exp_main_performance_profiles}) to provide a statistically robust comparison of overall efficiency and reliability. To further examine fine-grained behavior, we provide detailed learning curves (\Figure~\ref{exp_main_learning_curve}) showcasing performance dynamics on individual tasks. Together, these metrics offer a comprehensive assessment of algorithmic behavior and performance.

\textbf{Superior Sample Efficiency and Robustness}~As shown in \Figure~\ref{exp_main_performance_profiles}, \textbf{\texttt{ROSER}} achieves a significant Pareto improvement over the vanilla SAC and its augmented variants. In the initial threshold range ($\tau < 0.2$), \textbf{\texttt{ROSER}} maintains a run fraction near $1.0$, whereas baseline SAC and SAC+\textbf{\texttt{R}} show immediate degradation. This indicates that \textbf{\texttt{ROSER}} effectively mitigates the risk of catastrophic failure during training. In the more challenging Manipulation tasks, \textbf{\texttt{ROSER}} retains a higher success density at high-performance thresholds ($\tau > 0.8$) compared to other variant. These results highlight that \textbf{\texttt{ROSER}} not only delivers better performance but also exhibits greater robustness across different tasks.

\textbf{The Necessity of Principled Coordination}~A critical observation is the performance gap between \textbf{\texttt{ROSER}} and the naive stack baseline: \textbf{\texttt{ROSER}} delivers a pronounced improvement on locomotion tasks and a modest gain on manipulation tasks, most evident in the higher $\tau$ regime. This provides empirical evidence that principled coordination is crucial for unlocking the full potential of RL systems.

\textbf{Generalization Across Environments}~These conclusions are further corroborated by the learning curves in \Figure~\ref{exp_main_learning_curve}. Empirically, \textbf{\texttt{ROSER}} distinguishes itself through both rapid initial convergence and a higher final performance plateau across a diverse task suite. While single-component variants (e.g., SAC+\textbf{\texttt{R}}) show inconsistent improvements depending on the task, \textbf{\texttt{ROSER}} provides a universally stable performance boost, highlighting that our framework is environment-agnostic and robust, making it a general-purpose solution for sample-efficient RL.

\subsection{Analysis}

In this section, we analyze the efficacy of \textbf{$\texttt{R}\star$} and \textbf{U2P}. Our objective is to determine whether their performance gains stem from localized module improvements or from their capacity to mitigate emergent challenges within multi-module integration.

\textbf{Experimental Setup}~To disentangle standalone benefits from integration-driven synergies, we evaluate the effectiveness of \textbf{$\texttt{R}\star$} and \textbf{U2P} at two levels of complexity: (i) \textbf{Isolated Evaluation}: We integrate the component into a vanilla SAC baseline (e.g., SAC+\textbf{\texttt{ER}}(\textbf{P}) vs. SAC+\textbf{\texttt{ER}}(\textbf{U2P})) to establish a performance floor in simplified contexts; (ii) \textbf{Synergy Evaluation}: We perform studies within the full \textbf{\texttt{ROSER}} framework (e.g., \textbf{\texttt{ROSER}}(\textbf{P}) vs. \textbf{\texttt{ROSER}}(\textbf{U2P})) to quantify their contributions under complex, integrated conditions. By contrasting their standalone gains with contributions under synergy contexts, we characterize the mechanism underlying the observed improvements. We evaluate these configurations across a representative subset of 6 tasks: \textit{dog-run}, \textit{humanoid-run}, \textit{h1-balance-simple}, \textit{turnfaucet}, \textit{key-turn-hard}, \textit{pen-twirl-hard}, running 5 seeds for each experiment.

\begin{table}[h]
\small
\centering
\caption{\textbf{Analysis on Component Synergy.} IQM with 95\% stratified bootstrap CI (in brackets).}
\label{tab_analysis}
\addtolength{\tabcolsep}{-2pt}
\begin{tabular}{lcccc}
\toprule
& \multicolumn{2}{c}{\textbf{Representation}} & \multicolumn{2}{c}{\textbf{Experience Replay}} \\
\cmidrule(lr){2-3} \cmidrule(lr){4-5}
\textbf{Framework} & \texttt{\textbf{\texttt{R}}} & \textbf{$\texttt{R}\star$} & \texttt{\textbf{\texttt{ER}}}(\textbf{P}) & \texttt{\textbf{\texttt{ER}}}(\textbf{U2P}) \\
\midrule
Vanilla SAC & \textbf{0.175} ~\textcolor{gray}{[0.143, 0.209]} & 0.160 ~\textcolor{gray}{[0.098, 0.273]} & \textbf{0.250} ~\textcolor{gray}{[0.182, 0.295]} & 0.165 ~\textcolor{gray}{[0.128, 0.229]} \\
\textbf{\texttt{ROSER}} (Ours) & 0.514 ~\textcolor{gray}{[0.482, 0.541]} & \textbf{0.657} ~\textcolor{gray}{[0.524, 0.754]} & 0.671 ~\textcolor{gray}{[0.604, 0.723]} & \textbf{0.713} ~\textcolor{gray}{[0.671, 0.748]} \\
\bottomrule
\end{tabular}
\end{table}
\textbf{Analysis on Stable Model-based Representation ($\texttt{R}\star$).}~As illustrated in \Table~\ref{tab_analysis}, \textbf{$\texttt{R}\star$} degrades performance (-0.015) in the simplified setting, yet yields a clear improvement (+0.143) when integrated into the \textbf{\texttt{ROSER}} framework. This reversal indicates that \textbf{$\texttt{R}\star$} acts as a system-level stabilizer addressing joint optimization coupling, rather than a standalone optimizer.

\textbf{Analysis on Uniform-to-Prioritized (U2P).}~As illustrated in \Table~\ref{tab_analysis}, while \textbf{U2P} degrades performance (-0.085) in the standalone SAC-\textbf{\texttt{ER}} baseline, it yields a performance boost (+0.042) within the \textbf{\texttt{ROSER}} framework. This discrepancy suggests that the efficacy of \textbf{U2P} transcends simple optimization of prioritized experience replay. Instead, its primary value lies in buffer-level regularization, which mitigates the compound non-stationarity emergent from multi-module integration.

\textbf{Summary of Experimental Findings}~
Experiments demonstrate that \textbf{\texttt{ROSER}} consistently improves sample efficiency across a diverse set of continuous-control tasks. These results validate the design principles derived from our initial investigation, confirming that systematic coordination is important to performance gains. Notably, unlike \textbf{\texttt{ROSER}}, the benefits of isolated modules are highly sensitive to task characteristics, suggesting that individual sample-efficiency-enhancing techniques may have inherent limitations, making it difficult for them to address tasks with specific challenges or requirements. Overall, these findings underscore the value of a holistic, system-level perspective on sample-efficiency improvements.

%% file: Section/9_Conclusion.tex
\section{Conclusion}
\label{sec_conclusion}

In this paper, we investigated the synergistic potential of combining diverse sample-efficiency-enhancing techniques in Reinforcement Learning. We reveal that individually successful enhancements exhibit surprisingly divergent behaviors when jointly integrated, implying that simply stacking these modules does not guarantee additive performance gains. Rather, their interactions are complex and demand principled coordination. Through extensive empirical analysis, we demonstrated that the robustness of individual modules and their mutual stability act as the dominant factors in determining the overall sample efficiency, leading us to formalize three design principles for synergistic coordination.

To evaluate the rationality of our summarized design principles, we instantiate the \textbf{\texttt{ROSER}} framework. By incorporating a Stable Model-based Representation (\textbf{$\texttt{R}\star$}) and an Uniform-to-Prioritized Replay (\textbf{U2P}), \textbf{\texttt{ROSER}} effectively mitigates non-stationarity and unlocks the joint potential of its constituent parts. Our results show that this integrated framework not only stabilizes the learning trajectory but also achieves superior performance in both locomotion and manipulation tasks. Ultimately, this work highlights the importance of analyzing and improving RL sample efficiency from a synergistic, multi-module perspective, which is key to developing more efficient autonomous agents.

\textbf{Limitations.}~
While our study shows the power of synergistic design in improving sample efficiency, several avenues remain for future exploration. First, although our empirical evidence shows that individual enhancement modules often exhibit significant task-dependency, a systematic taxonomy of which environment features, such as reward density or state space complexity, favor specific techniques is yet to be established. Second, our integration strategies, such as \textbf{$\texttt{R}\star$} and the \textbf{U2P}, are primarily empirically-grounded; future research could focus on discovering more theoretically optimal coordination mechanisms. Furthermore, our study focused on three specific components, yet other crucial factors like exploration could be integrated to further expand the synergistic potential. Finally, although our findings are robust under the SAC baseline, extending this investigation to other off-policy algorithms, as well as more diverse real-world benchmarks, will be essential to establish broader generalizability.

%% file: Section/Appendix.tex
\begin{center}
	{\LARGE \bf Appendix}
\end{center}
\bigskip

\hrule
\vskip 0.2in
\startcontents[sections]
\printcontents[sections]{}{1}{\setcounter{tocdepth}{3}}
\vskip 0.2in
\hrule

\input{Appendix/DDPG}
\input{Appendix/Hyperparameters}

\input{Appendix/Architecture}
\input{Appendix/Environments}
\input{Appendix/Analysis}
\input{Appendix/Results}

%% file: Appendix/DDPG.tex
\newpage
\section{\textbf{\texttt{ROSER}} on DDPG}
\label{appendix_ddpg}

\begin{figure}[ht]
\centering
\vspace{-0.6\baselineskip}
\begin{center}
\includegraphics[width=0.8\linewidth]{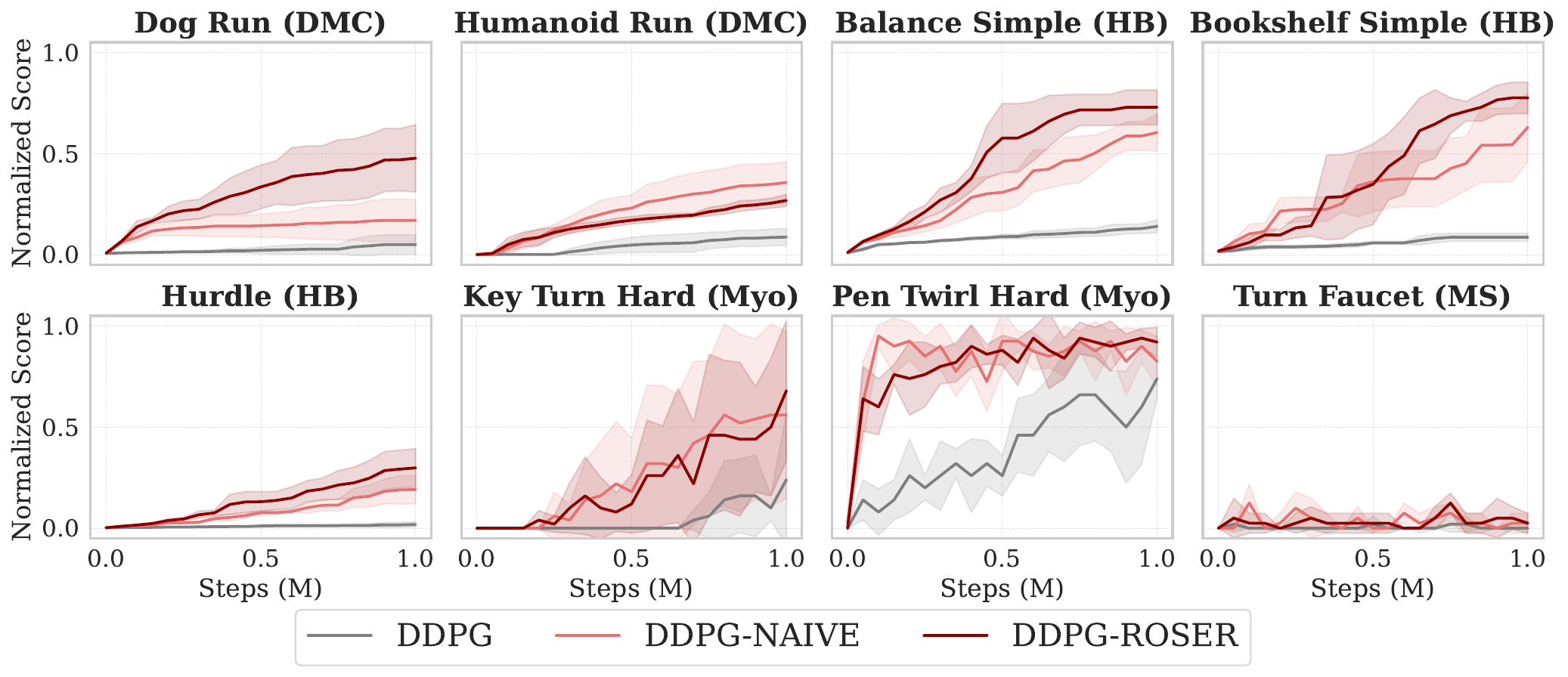}
\end{center}
\caption{
  \textbf{Learning curves.} Validating the Generalizability of \textbf{\texttt{ROSER}} on the DDPG Algorithm.
}
\vspace{-0.8\baselineskip}
\label{ddpg_learning_curves_appendix}
\end{figure}

\begin{table}[htbp]
\centering
\caption{\textbf{Performance comparison of DDPG variants.} Values represent Median, IQM, and Mean with 95\% confidence intervals.}
\label{tab:ddpg_comparison}
\footnotesize
\addtolength{\tabcolsep}{-2pt}
\begin{tabular}{l lll lll}
\toprule
& \multicolumn{3}{c}{\textbf{Locomotion}} & \multicolumn{3}{c}{\textbf{Manipulation}} \\
\cmidrule(r){2-4} \cmidrule(l){5-7}
\textbf{Algorithm} & Median & IQM & Mean & Median & IQM & Mean \\
\midrule
DDPG (Vanilla) & 0.0683 & 0.0683 & 0.0738 & 0.2030 & 0.1913 & 0.3215 \\
& {\scriptsize [0.0416, 0.0969]} & {\scriptsize [0.0452, 0.0934]} & {\scriptsize [0.0579, 0.0906]} & {\scriptsize [0.0818, 0.3700]} & {\scriptsize [0.0981, 0.3196]} & {\scriptsize [0.2443, 0.4089]} \\
\addlinespace
DDPG+Naive Stack & 0.2741 & 0.3030 & 0.3302 & 0.6249 & 0.6549 & 0.6025 \\
& {\scriptsize [0.2260, 0.3382]} & {\scriptsize [0.2390, 0.3667]} & {\scriptsize [0.2899, 0.3734]} & {\scriptsize [0.4075, 0.8373]} & {\scriptsize [0.4504, 0.8373]} & {\scriptsize [0.4903, 0.7091]} \\
\addlinespace
DDPG+ROSER & \textbf{0.3876} & \textbf{0.3964} & \textbf{0.4429} & \textbf{0.7377} & \textbf{0.7577} & \textbf{0.6589} \\
& {\scriptsize \textbf{[0.3132, 0.4735]}} & {\scriptsize \textbf{[0.3412, 0.4650]}} & {\scriptsize \textbf{[0.3954, 0.4898]}} & {\scriptsize \textbf{[0.5577, 0.8616]}} & {\scriptsize \textbf{[0.6131, 0.8616]}} & {\scriptsize \textbf{[0.5686, 0.7223]}} \\
\bottomrule
\end{tabular}
\end{table}

To verify that \textbf{\texttt{ROSER}} is algorithm-agnostic, we conducted additional experiments using DDPG on 8 tasks (\textit{dog-run}, \textit{humanoid-run}, \textit{h1-balance-simple}, \textit{h1-hurdle}, \textit{h1-bookshelf-simple}, \textit{turnfaucet}, \textit{key-turn-hard} and \textit{pen-twirl-hard}) with 5 seeds. The results, shown in the \Figure~\ref{ddpg_learning_curves_appendix} and \Table~\ref{tab:ddpg_comparison}, confirm that \textbf{\texttt{ROSER}} improves DDPG performance over both the vanilla version and naive stacking.

%% file: Appendix/Hyperparameters.tex
\newpage
\section{Hyperparameters}
\label{appendix_hyperparameters}

\begin{table}[ht]
\footnotesize
\centering
\parbox{\textwidth}{
\caption{\textbf{Backbone hyperparameters.} We provide a detailed list of the hyperparameters used in the backbone (SAC) and training setup. We automatically set the discount factor based on the heuristics from TD-MPC2~\cite{hansen2024tdmpc2}.}
\label{appendix:backbone_hparams}
\centering
\begin{tabular}{lr}
\toprule
\textbf{Hyperparameter}         & \textbf{Value} \\ \midrule
Critic block type               & MLP \\
Critic hidden dim               & 512 \\
Critic learning rate            & 1e-4 \\
Critic activation function      & ReLU \\
Target critic momentum ($\tau$) & 5e-3 \\
Clipped Double Q                & True \\ \midrule
Actor block type                & MLP \\
Actor hidden dim                & 128  \\
Actor learning rate             & 1e-4 \\
Actor activation function       & ReLU \\ \midrule
Initial temperature ($\alpha_0$) & 1e-2 \\
Temperature learning rate       & 1e-4 \\
Target entropy ($\mathcal{H}^*$) & $|\mathcal{A}|/2$ \\ \midrule
Optimizer                       & AdamW \\
Optimizer momentum ($\beta_1$, $\beta_2$) & (0.9, 0.999) \\
Weight decay ($\lambda$)        & 1e-2 \\ \midrule
Discount ($\gamma$)             & Heuristic \\
Replay ratio                    & 2 \\
Multi-step returns horizon      & 1 \\
\bottomrule
\end{tabular}}
\end{table}

\begin{table}[H]
\footnotesize
\centering
\parbox{\textwidth}{
\caption{\textbf{\texttt{R} hyperparameters.} We provide a detailed list of the hyperparameters used in model-based representation. We follow the hyperparameters specified in the original MR.Q paper~\cite{fujimoto2025towards}. The encoder architecture is conditional: SimBa Residual (w/ \textbf{\texttt{OS}}) vs. MLP (w/o \textbf{\texttt{OS}}).} 
\label{appendix:R_hparams}
\centering
\begin{tabular}{lr}
\toprule
\textbf{Hyperparameter}         & \textbf{Value} \\ \midrule
$\mathbf{z}_s$ dim              & 512 \\
$\mathbf{z}_{sa}$ dim           & 256 \\
$\mathbf{z}_a$ dim              & 512 \\
Encoder horizon                 & 5 \\
Encoder block type              & MLP/SimBa Residual \\
Encoder num blocks              & 1 \\
Encoder hidden dim              & 512  \\
Encoder learning rate           & 1e-4 \\
Encoder activation function     & ELU \\
Encoder update frequency        & 250 \\
Optimizer                       & AdamW \\
Optimizer momentum ($\beta_1$, $\beta_2$) & (0.9, 0.999) \\
Weight decay ($\lambda$)        & 1e-4 \\
Dynamics loss weight            & 1 \\
Reward loss weight              & 0.1 \\
Terminal loss weight            & 0.1 \\
Reward bins                     & 65 \\
Reward range                    & $[-10,10]$ \\
Replay ratio                    & 1 \\
\bottomrule
\end{tabular}
}
\end{table}

\begin{table}[ht]
\footnotesize
\centering
\parbox{\textwidth}{
\caption{\textbf{\texttt{OS} hyperparameters.} We provide a detailed list of the hyperparameters used in optimization stability. We follow the hyperparameters specified in the original SimBa paper~\cite{lee2025simba}.}
\label{appendix:OS_hparams}
\centering
\begin{tabular}{lr}
\toprule
\textbf{Hyperparameter}         & \textbf{Value} \\ \midrule
Critic block type               & SimBa Residual \\
Critic num blocks               & 2 \\
Critic hidden dim               & 512 \\
Critic learning rate            & 1e-4 \\
Critic activation function      & ReLU \\
Target critic momentum ($\tau$) & 5e-3 \\
Clipped Double Q                & True \\ \midrule
Actor block type                & SimBa Residual \\
Actor num blocks                & 1 \\
Actor hidden dim                & 128  \\
Actor learning rate             & 1e-4 \\
Actor activation function       & ReLU \\
\bottomrule
\end{tabular}
}
\vspace{0.15in}
\end{table}

\begin{table}[ht]
\footnotesize
\centering
\parbox{\textwidth}{
\caption{\textbf{\texttt{ER} hyperparameters.} We provide a detailed list of the hyperparameters used in experience replay. We follow the hyperparameters specified in the original ReLo paper~\cite{sujit2022prioritizing}. Following ~\cite{fujimoto2020equivalence}, when in the uniform sampling state (i.e., $\alpha=0$), the critic utilizes mean squared error (MSE) loss. In contrast, when in a non-uniform sampling state (i.e., $\alpha>0$), the critic adopts Huber loss.}
\label{appendix:ER_hparams}
\centering
\begin{tabular}{lr}
\toprule
\textbf{Hyperparameter}         & \textbf{Value} \\ \midrule
Buffer size                     & 1M \\
Batch size                      & 256 \\ 
ReLo Clipping Offset $\epsilon$ & 0.01 \\ \midrule
$\alpha_F$                      & 0.4 \\
$t_{\text {start }}$            & 0.2 * Total env. steps \\
$t_{\text {end }}$              & 0.8 * Total env. steps \\
Critic Loss                     & Huber/MSE \\
\bottomrule
\end{tabular}}
\end{table}

%% file: Appendix/Architecture.tex
\newpage
\section{Architecture}
\label{appendix_architecture}
This section outlines the networks employed in our paper, including MLP Block, Simba Block, Encoder, implemented using PyTorch2~\cite{Ansel_PyTorch_2_Faster_2024}. The implementation of the Critic and Actor follows~\cite{lee2025simba}, therefore, we do not provide further details here.

\subsection{MLP Block}
The MLP Block is a basic fully connected neural network module consisting of two linear layers with non-linear activation function applied behind them.
\begin{lstlisting}
import math
import torch
import torch.nn as nn

class MLPBlock(nn.Module):
    def __init__(
self,
        input_dim: int,
        hidden_dim: int,
        dtype: torch.dtype,
        activ='ReLU'
     ):
        super().__init__()

        self.fc1 = nn.Linear(input_dim, hidden_dim, dtype=dtype)
        self.fc2 = nn.Linear(hidden_dim, hidden_dim, dtype=dtype)
        self.activ1 = getattr(nn, activ)()
        self.activ2 = getattr(nn, activ)()

        orthogonal_init_(self.fc1, gain=math.sqrt(2))
        orthogonal_init_(self.fc2, gain=math.sqrt(2))

     def forward(self, x: torch.Tensor) -> torch.Tensor:
        x = self.fc1(x)
        x = self.activ1(x)
        x = self.fc2(x)
        x = self.activ2(x)

        return x
\end{lstlisting}

\subsection{Simba Block}
\label{appendix_simba_block}
The Simba Block consists of a series of residual blocks that allow for more complex transformations while retaining input information through shortcut connections. This block enables deeper networks by maintaining stable gradient flow during training. It also employs layer normalization for stability. Our implementation follows~\cite{lee2025simba}.
\begin{lstlisting}
class ResidualBlock(nn.Module):
    def __init__(
        self,
        hidden_dim: int,
        dtype: torch.dtype,
        activ='ReLU'
    ):
        super().__init__()

        self.layer_norm = nn.LayerNorm(hidden_dim, dtype=dtype)
        self.fc1 = nn.Linear(hidden_dim, hidden_dim*4, dtype=dtype)
        self.fc2 = nn.Linear(hidden_dim*4, hidden_dim, dtype=dtype)
        self.activ = getattr(nn, activ)()

        he_normal_init_(self.fc1)
        he_normal_init_(self.fc2)
    
    def forward(self, x: torch.Tensor) -> torch.Tensor:
        res = x
        x = self.layer_norm(x)
        x = self.fc1(x)
        x = self.activ(x)
        x = self.fc2(x)
        
        return res + x

class SimbaBlock(nn.Module):
    def __init__(
        self,
        num_blocks: int,
        input_dim: int,
        hidden_dim: int,
        dtype: torch.dtype,
        activ='ReLU'
    ):
        super().__init__()

        self.fc = nn.Linear(input_dim, hidden_dim, dtype=dtype)
        self.residual_blocks = nn.ModuleList([
            ResidualBlock(hidden_dim, dtype=dtype, activ=activ) for _ in range(num_blocks)
        ])
        self.layer_norm = nn.LayerNorm(hidden_dim, dtype=dtype)

        orthogonal_init_(self.fc)

    def forward(self, x: torch.Tensor) -> torch.Tensor:
        x = self.fc(x)
        for block in self.residual_blocks:
            x = block(x)
        x = self.layer_norm(x)
        
        return x
\end{lstlisting}

\subsection{Encoder}
\label{appendix_encoder}
The Encoder combines the state encoder $Z_s$ and state-action encoder $Z_{sa}$ to map the input state and action data into latent representations, which are then used for further processing in the reinforcement learning pipeline. Our implementation follows~\cite{fujimoto2025towards}.
\begin{lstlisting}
class ROSEREncoder(nn.Module):
    def __init__(
        self,
        state_dim: int,
        hidden_dim: int,
        action_dim: int,
        zs_dim: int,
        za_dim: int,
        zsa_dim: int,
        num_bins: int,
        num_blocks: int,
        dtype: torch.dtype,
        activ='ELU'
    ):
        super().__init__()
        self.zs_dim = zs_dim
        
        self.zs = Embedding(
            num_blocks=num_blocks,
            input_dim=state_dim,
            hidden_dim=hidden_dim,
            output_dim=zs_dim,
            dtype=dtype,
            activ='ELU'
        )
        self.za = self.mlp_za
        self.fc = nn.Linear(action_dim, za_dim, dtype=dtype)
        self.zsa = Embedding(
            num_blocks=num_blocks,
            input_dim=zs_dim+za_dim,
            hidden_dim=hidden_dim,
            output_dim=zsa_dim,
            dtype=dtype,
            activ='ELU'
        )
        self.model = nn.Linear(zsa_dim, num_bins + zs_dim + 1)

        self.activ = getattr(nn, activ)()

        orthogonal_init_(self.model)
\end{lstlisting}
It is important to note that in the encoder implementation described above, both the state encoder $Z_s$ and the state-action encoder $Z_{sa}$ are instances of the \textit{Embedding} object. In the case of experiments solely on \textbf{\texttt{R}}, the embedding follows the original MR.Q. However, in scenarios where \textbf{\texttt{R}} is combined with \textbf{\texttt{OS}}, the embedding is implemented as follows, inheriting the structure from the Simba Block.
\begin{lstlisting}
class Embedding(SimbaBlock):
    def __init__(
        self,
        num_blocks: int,
        input_dim: int,
        hidden_dim: int,
        output_dim: int,
        dtype: torch.dtype,
        activ='ELU'
    ):
        super().__init__(
            num_blocks=num_blocks,
            input_dim=input_dim,
            hidden_dim=hidden_dim,
            dtype=dtype,
            activ=activ
        )
        self.head = nn.Linear(hidden_dim, output_dim, dtype=dtype)

        orthogonal_init_(self.head)

    def forward(self, x: torch.Tensor) -> torch.Tensor:
        x = super().forward(x)
        x = self.head(x)
        return x
\end{lstlisting}

%% file: Appendix/Environments.tex
\newpage
\section{Environments}
\label{appendix_environments}

\textbf{DeepMind Control suite}~\cite{tassa2018dmc} is a standardized benchmark test set for reinforcement learning developed by DeepMind based on the MuJoCo physics engine. It focuses on continuous control tasks, covering a wide range of motion capture and motion control scenarios from simple handstands to high-dimensional humanoid robots. This suite is renowned for its extremely high code quality, unified reward mechanism, and outstanding support for pixel-level observation (Visual RL), and is one of the most commonly used performance evaluation criteria in papers in fields such as robot control, representation learning, and model predictive control. We evaluate 2 DMC-Hard tasks, with details provided in \Table~\ref{appendix:dmc}.

\textbf{HumanoidBench}~\cite{sferrazza2024humanoidbench} is a benchmark test specifically designed for humanoid robots, used to evaluate and compare their movement and task execution capabilities in complex environments. It challenges the robot's performance in gait control, balance, object grasping and other aspects through a series of highly simulated tasks. This benchmark test utilizes a high-quality physical simulation environment to ensure the authenticity and diversity of tasks and actions, making it an important evaluation tool in fields such as reinforcement learning and robot control. We consider 11 tasks, with details provided in \Table~\ref{appendix:hb}.

\textbf{MyoSuite}~\cite{MyoSuite2022} is an integrated platform that combines musculoskeletal simulation with artificial intelligence, aiming to provide standardized benchmarks for studying the evolution and learning of physiological and neuro-motor control by establishing highly accurate biomechanical models in physiology. This benchmark suite not only covers multi-dimensional task challenges ranging from tendon transfer surgery modeling to complex object manipulation, but also bridges the gap between neural function recovery research and modern reinforcement learning algorithms. We focus on 2 tasks, with details provided in \Table~\ref{appendix:myo}

\textbf{Maniskill2}~\cite{gu2023maniskill2} is a high-quality benchmark for learning robot operation skills, aiming to promote the development of reinforcement learning and imitation learning fields. This benchmark offers a lot of challenging task categories, covering various robot operation tasks in real-world scenarios, such as grasping, manipulating, and assembling. ManiSkill2 supports over 2,000 different object models and more than 4 million frames of demonstration data, providing diverse and high-quality training and testing resources. Through deep integration with the SAPIEN engine, ManiSkill2 not only supports high-speed visual input but also features flexible environment configuration, allowing researchers to conduct efficient algorithm evaluations in standardized environments. The openness and efficiency of this benchmark make it an ideal testing platform for robot control and learning algorithms, effectively promoting the comparison and progress of various learning methods. We select 3 tasks, with details provided in \Table~\ref{appendix:maniskill}

\begin{table}[ht]
\footnotesize
\centering
\parbox{\textwidth}{
\caption{\textbf{Environment details.} We list the episode length, action repeat for each domain, total environment steps, and performance metrics used for our experiments. It is worth noting that, unless otherwise stated, for tasks that use Return as the performance metric, the normalized score is calculated using 1000 as the normalization factor (10000 for Reach in HumanoidBench, as the return for this task greatly exceeds 1000).}
\label{appendix:environment_detail}
\centering
\begin{tabular}{lcccc}
\toprule
 & \textbf{DMC} & \textbf{HumanoidBench} & \textbf{MyoSuite}  & \textbf{Maniskill} \\ \midrule
Episode length & 1,000 & $500-1,000$ & 100 & 200 \\
Action repeat & 2 & 2 & 2 & 2 \\
Effective length & 500 & $250-500$ & 50 & 100 \\
Total env. steps & 1 M & 1 M & 1 M & 1 M \\
Performance metric & Return & Return & Success Rate & Success Rate \\
\bottomrule
\end{tabular}
}
\vspace{0.15in}
\end{table}

\begin{table}[ht]
\footnotesize
\centering
\parbox{\textwidth}{
\caption{\textbf{DMC.} We consider 2 DMC-Hard tasks, both of which are locomotion tasks, as shown below.}
\label{appendix:dmc}
\centering
\begin{tabular}{lccc}
\toprule
\textbf{Task} & \textbf{Observation dim} & \textbf{Action dim} & \textbf{Task Category}\\ \midrule
Dog Run & 223 & 38 & loco\\
Humanoid Run & 67 & 24 & loco\\
\bottomrule
\end{tabular}
}
\end{table}

\begin{table}[ht]
\footnotesize
\centering
\parbox{\textwidth}{
\caption{\textbf{HumanoidBench.} We consider 11 HumanoidBench tasks, as shown below, with the first 7 tasks being locomotion tasks and the last 4 tasks being manipulation tasks.}
\label{appendix:hb}
\centering
\begin{tabular}{lccc}
\toprule
\textbf{Task} & \textbf{Observation dim} & \textbf{Action dim} & \textbf{Task Category}\\ \midrule
Balance Simple      & 64    & 19    & loco\\
Balance Hard        & 77    & 19    & loco\\
Run                 & 51    & 19    & loco\\
Hurdle              & 51    & 19    & loco\\
Reach               & 57    & 19    & loco\\
Sit Hard            & 64    & 19    & loco\\
Slide               & 51    & 19    & loco\\ \midrule
Bookshelf Simple    & 308   & 61    & mani\\
Basketball          & 64    & 19    & mani\\
Door                & 55    & 19    & mani\\
Spoon               & 167   & 61    & mani\\
\bottomrule
\end{tabular}
}
\end{table}

\begin{table}[ht]
\footnotesize
\centering
\parbox{\textwidth}{
\caption{\textbf{MyoSuite.} We consider 2 Myosuite tasks, both of which are manipulation tasks, as shown below.}
\label{appendix:myo}
\centering
\begin{tabular}{lccc}
\toprule
\textbf{Task} & \textbf{Observation dim} & \textbf{Action dim} & \textbf{Task Category} \\ \midrule
Key Turn Hard       & 93 & 39 & mani\\
Pen Twirl Hard      & 83 & 39 & mani\\
\bottomrule
\end{tabular}
}
\end{table}

\begin{table}[H]
\footnotesize
\centering
\parbox{\textwidth}{
\caption{\textbf{Maniskill.} We consider 3 Maniskill tasks, all of which are manipulation tasks, as shown below.}
\label{appendix:maniskill}
\centering
\begin{tabular}{lccc}
\toprule
\textbf{Task} & \textbf{Observation dim} & \textbf{Action dim} & \textbf{Task Category} \\ \midrule
LiftCube        & 42 & 4 & mani\\
PickCube        & 51 & 4 & mani\\
TurnFaucet      & 40 & 7 & mani\\
\bottomrule
\end{tabular}
}
\end{table}

%% file: Appendix/Analysis.tex
\newpage
\section{Analysis of \texttt{ER}}
\label{appendix_analysis}

\begin{figure}[ht]
  \vskip 0.2in
  \begin{center}
    \centerline{\includegraphics[width=0.5\columnwidth]{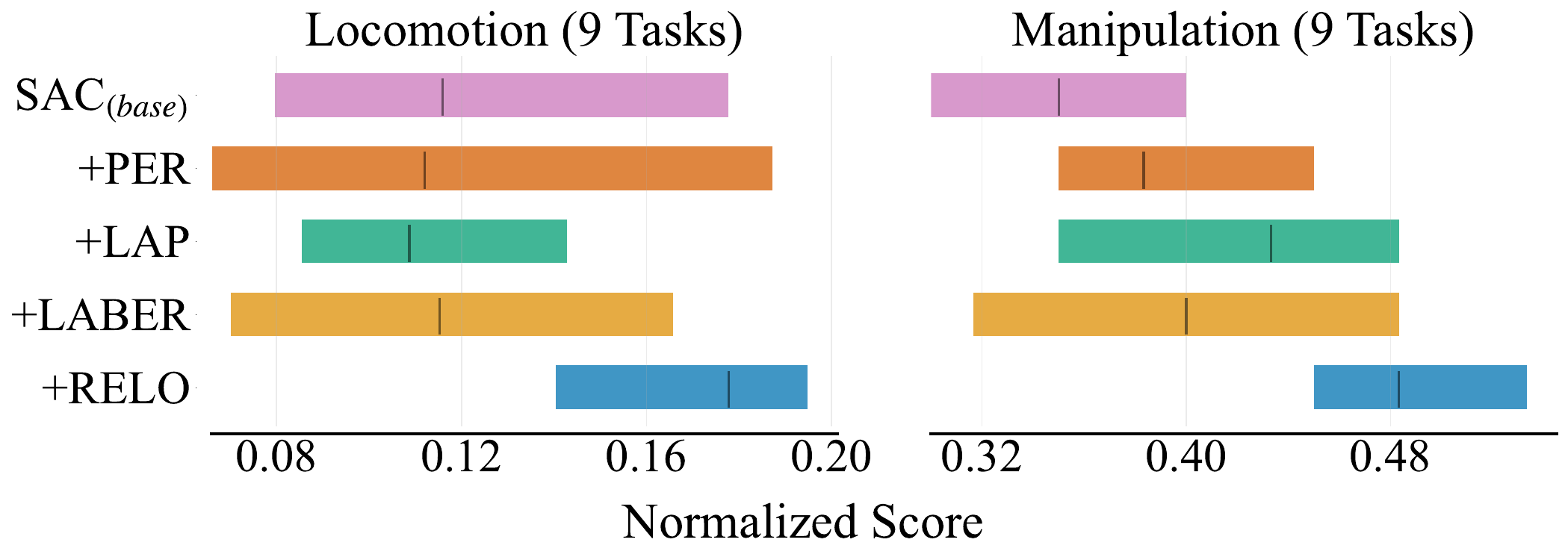}}
    \caption{
      \textbf{Comparision of Different Priority Experience Replay Design.} 
    }
    \label{exp_ab_experience_replay_design}
  \end{center}
\end{figure}

We next study the impact of different experience replay strategies on sample efficiency.
To this end, we compare several representative replay schemes, including Uniform Replay, Prioritized Experience Replay (\textbf{PER}~\cite{schaul2015prioritized}), Loss-Adjusted Prioritized experience replay (\textbf{LAP}~\cite{fujimoto2020equivalence}), Large Batch Experience Replay (\textbf{LaBER}~\cite{lahire2021large}), and prioritized experience replay based on Reducible Loss (\textbf{ReLo}~\cite{sujit2022prioritizing}).
These methods span a range of prioritization mechanisms, from uniform sampling to TD-error-based and loss-aware criteria. We assess these methods on a representative set of four tasks: \textit{dog-run}, \textit{h1-balance-simple}, \textit{turnfaucet}, and \textit{pen-twirl-hard}, conducting 5 seeds per experiment.

Results shown in \cref{exp_ab_experience_replay_design} reveal that other prioritized experience replay methods perform moderately on the locomotion task, with only a notable improvement on the manipulation task. In contrast, ReLo-based prioritized replay demonstrates significant improvement across both task types, consistently outperforming all other methods by a large margin.
Overall, this analysis supports the rationale for using ReLo-based prioritized replay as the representative technique for \textbf{\texttt{ER}} in our investigation.

%% file: Appendix/Results.tex
\newpage
\section{Complete Main Results}
\label{appendix_results}

\begin{table}[ht]
\footnotesize
\addtolength{\tabcolsep}{-5pt}
\newcolumntype{Y}{>{\centering\arraybackslash}X} %
\caption{\textbf{Complete Main Results} Average performance across eight independent seeds, evaluated at 500k training steps (equivalent to 1M environment steps with an action repeat of 2). \tabledescript default reward. The best performance for each task is highlighted in bold.}
\begin{tabularx}{\textwidth}{l@{\hspace{4pt}} r@{}X@{\hspace{0pt}} r@{}X@{\hspace{4pt}} r@{}X@{\hspace{4pt}} r@{}X@{\hspace{4pt}} r@{}X@{\hspace{4pt}} r@{}l@{}}
\toprule
Task & & \multicolumn{1}{l}{SAC} & & \multicolumn{1}{l}{SAC+\textbf{\texttt{R}}} & & \multicolumn{1}{l}{SAC+\textbf{\texttt{OS}}} & & \multicolumn{1}{l}{SAC+\textbf{\texttt{ER}}} & & \multicolumn{1}{l}{Naive Stack} & & \multicolumn{1}{l}{\textbf{\texttt{ROSER}}} \\

\midrule
Dog Run & 49 & ~\textcolor{gray}{\scriptsize[11, 98]} & 12 & ~\textcolor{gray}{\scriptsize[10, 15]} & 558 & ~\textcolor{gray}{\scriptsize[500, 599]} & 141 & ~\textcolor{gray}{\scriptsize[82, 189]} & 229 & ~\textcolor{gray}{\scriptsize[149, 333]} & \textbf{632} & ~\textcolor{gray}{\scriptsize[525, 723]} \\
Humanoid Run & 119 & ~\textcolor{gray}{\scriptsize[84, 140]} & 115 & ~\textcolor{gray}{\scriptsize[48, 172]} & 171 & ~\textcolor{gray}{\scriptsize[159, 183]} & 133 & ~\textcolor{gray}{\scriptsize[120, 147]} & \textbf{406} & ~\textcolor{gray}{\scriptsize[359, 449]} & 380 & ~\textcolor{gray}{\scriptsize[315, 438]} \\
Run & 164 & ~\textcolor{gray}{\scriptsize[54, 305]} & 45 & ~\textcolor{gray}{\scriptsize[27, 65]} & 249 & ~\textcolor{gray}{\scriptsize[189, 346]} & 92 & ~\textcolor{gray}{\scriptsize[59, 143]} & \textbf{822} & ~\textcolor{gray}{\scriptsize[818, 825]} & \textbf{822} & ~\textcolor{gray}{\scriptsize[818, 825]} \\
Balance Simple & 195 & ~\textcolor{gray}{\scriptsize[165, 232]} & 472 & ~\textcolor{gray}{\scriptsize[349, 590]} & 266 & ~\textcolor{gray}{\scriptsize[177, 370]} & 178 & ~\textcolor{gray}{\scriptsize[164, 192]} & 772 & ~\textcolor{gray}{\scriptsize[725, 815]} & \textbf{825} & ~\textcolor{gray}{\scriptsize[809, 838]} \\
Balance Hard & 63 & ~\textcolor{gray}{\scriptsize[59, 68]} & 79 & ~\textcolor{gray}{\scriptsize[67, 92]} & 79 & ~\textcolor{gray}{\scriptsize[72, 86]} & 64 & ~\textcolor{gray}{\scriptsize[60, 69]} & 106 & ~\textcolor{gray}{\scriptsize[93, 119]} & \textbf{147} & ~\textcolor{gray}{\scriptsize[128, 170]} \\
Sit Hard & 496 & ~\textcolor{gray}{\scriptsize[268, 717]} & 35 & ~\textcolor{gray}{\scriptsize[9, 85]} & 613 & ~\textcolor{gray}{\scriptsize[463, 766]} & 550 & ~\textcolor{gray}{\scriptsize[356, 717]} & 841 & ~\textcolor{gray}{\scriptsize[836, 846]} & \textbf{853} & ~\textcolor{gray}{\scriptsize[846, 861]} \\
Reach & 3510 &\textcolor{gray}{\scriptsize[3226, 3810]} & 1104 &\textcolor{gray}{\scriptsize[957, 1273]} & 4402 &\textcolor{gray}{\scriptsize[3863, 4953]} & 4435 &\textcolor{gray}{\scriptsize[4003, 4794]} & 6168 &\textcolor{gray}{\scriptsize[4951, 7330]} & \textbf{7141} & ~\textcolor{gray}{\scriptsize[6574, 7664]} \\
Hurdle & 42 & ~\textcolor{gray}{\scriptsize[17, 67]} & 106 & ~\textcolor{gray}{\scriptsize[74, 134]} & 212 & ~\textcolor{gray}{\scriptsize[199, 226]} & 68 & ~\textcolor{gray}{\scriptsize[49, 83]} & 283 & ~\textcolor{gray}{\scriptsize[220, 335]} & \textbf{340} & ~\textcolor{gray}{\scriptsize[321, 357]} \\
Slide & 183 & ~\textcolor{gray}{\scriptsize[125, 235]} & 389 & ~\textcolor{gray}{\scriptsize[308, 476]} & 261 & ~\textcolor{gray}{\scriptsize[213, 298]} & 182 & ~\textcolor{gray}{\scriptsize[134, 223]} & 455 & ~\textcolor{gray}{\scriptsize[418, 513]} & \textbf{458} & ~\textcolor{gray}{\scriptsize[439, 478]} \\
Bookshelf Simple & 99 & ~\textcolor{gray}{\scriptsize[61, 145]} & 52 & ~\textcolor{gray}{\scriptsize[41, 64]} & 718 & ~\textcolor{gray}{\scriptsize[695, 739]} & 112 & ~\textcolor{gray}{\scriptsize[90, 135]} & 729 & ~\textcolor{gray}{\scriptsize[609, 823]} & \textbf{812} & ~\textcolor{gray}{\scriptsize[800, 824]} \\
Basketball & 80 & ~\textcolor{gray}{\scriptsize[44, 128]} & 28 & ~\textcolor{gray}{\scriptsize[23, 33]} & 163 & ~\textcolor{gray}{\scriptsize[102, 228]} & 97 & ~\textcolor{gray}{\scriptsize[62, 133]} & \textbf{353} & ~\textcolor{gray}{\scriptsize[306, 400]} & 306 & ~\textcolor{gray}{\scriptsize[222, 380]} \\
Door & 180 & ~\textcolor{gray}{\scriptsize[119, 232]} & 172 & ~\textcolor{gray}{\scriptsize[73, 272]} & 296 & ~\textcolor{gray}{\scriptsize[287, 308]} & 139 & ~\textcolor{gray}{\scriptsize[74, 205]} & 323 & ~\textcolor{gray}{\scriptsize[310, 338]} & \textbf{341} & ~\textcolor{gray}{\scriptsize[332, 348]} \\
Spoon & 19 & ~\textcolor{gray}{\scriptsize[16, 22]} & 23 & ~\textcolor{gray}{\scriptsize[13, 41]} & \textbf{380} & ~\textcolor{gray}{\scriptsize[376, 384]} & 26 & ~\textcolor{gray}{\scriptsize[22, 29]} & 350 & ~\textcolor{gray}{\scriptsize[338, 363]} & 369 & ~\textcolor{gray}{\scriptsize[351, 384]} \\
Key Turn Hard & 0.25 & ~\textcolor{gray}{\scriptsize[0.08, 0.50]} & 0.08 & ~\textcolor{gray}{\scriptsize[0.01, 0.15]} & 0.50 & ~\textcolor{gray}{\scriptsize[0.21, 0.79]} & 0.64 & ~\textcolor{gray}{\scriptsize[0.43, 0.80]} & 0.86 & ~\textcolor{gray}{\scriptsize[0.74, 0.98]} & \textbf{0.95} & ~\textcolor{gray}{\scriptsize[0.88, 1.00]} \\
Pen Twirl Hard & 0.68 & ~\textcolor{gray}{\scriptsize[0.59, 0.76]} & 0.98 & ~\textcolor{gray}{\scriptsize[0.94, 1.00]} & 0.94 & ~\textcolor{gray}{\scriptsize[0.88, 0.99]} & 0.74 & ~\textcolor{gray}{\scriptsize[0.65, 0.81]} & \textbf{1.00} & ~\textcolor{gray}{\scriptsize[1.00, 1.00]} & \textbf{1.00} & ~\textcolor{gray}{\scriptsize[1.00, 1.00]} \\
PickCube & 0 & ~\textcolor{gray}{\scriptsize[0, 0]} & 0 & ~\textcolor{gray}{\scriptsize[0, 0]} & 0 & ~\textcolor{gray}{\scriptsize[0, 0]} & 0.01 & ~\textcolor{gray}{\scriptsize[0, 0.04]} & \textbf{0.30} & ~\textcolor{gray}{\scriptsize[0.25, 0.35]} & 0.29 & ~\textcolor{gray}{\scriptsize[0.20, 0.38]} \\
TurnFaucet & 0.09 & ~\textcolor{gray}{\scriptsize[0.06, 0.10]} & 0.04 & ~\textcolor{gray}{\scriptsize[0.01, 0.08]} & 0.11 & ~\textcolor{gray}{\scriptsize[0.08, 0.15]} & 0.19 & ~\textcolor{gray}{\scriptsize[0.15, 0.23]} & \textbf{0.20} & ~\textcolor{gray}{\scriptsize[0.16, 0.24]} & \textbf{0.20} & ~\textcolor{gray}{\scriptsize[0.15, 0.25]} \\
LiftCube & 0.03 & ~\textcolor{gray}{\scriptsize[0.00, 0.06]} & 0.65 & ~\textcolor{gray}{\scriptsize[0.46, 0.83]} & 0.01 & ~\textcolor{gray}{\scriptsize[0.00, 0.04]} & 0.18 & ~\textcolor{gray}{\scriptsize[0.08, 0.26]} & \textbf{1.00} & ~\textcolor{gray}{\scriptsize[1.00, 1.00]} & \textbf{1.00} & ~\textcolor{gray}{\scriptsize[1.00, 1.00]} \\

\midrule
Mean & 289 & ~\textcolor{gray}{\scriptsize[266, 312]} & 146 & ~\textcolor{gray}{\scriptsize[132, 161]} & 465 & ~\textcolor{gray}{\scriptsize[432, 498]} & 345 & ~\textcolor{gray}{\scriptsize[319, 369]} & 658 & ~\textcolor{gray}{\scriptsize[589, 722]} & \textbf{746} & ~\textcolor{gray}{\scriptsize[713, 776]} \\
Median & 72 & ~\textcolor{gray}{\scriptsize[54, 91]} & 40 & ~\textcolor{gray}{\scriptsize[30, 55]} & 230 & ~\textcolor{gray}{\scriptsize[195, 247]} & 95 & ~\textcolor{gray}{\scriptsize[71, 112]} & 337 & ~\textcolor{gray}{\scriptsize[315, 346]} & \textbf{355} & ~\textcolor{gray}{\scriptsize[338, 373]} \\
IQM & 60 & ~\textcolor{gray}{\scriptsize[49, 72]} & 37 & ~\textcolor{gray}{\scriptsize[29, 47]} & 203 & ~\textcolor{gray}{\scriptsize[191, 216]} & 76 & ~\textcolor{gray}{\scriptsize[66, 87]} & 310 & ~\textcolor{gray}{\scriptsize[293, 329]} & \textbf{374} & ~\textcolor{gray}{\scriptsize[357, 390]} \\

\bottomrule
\end{tabularx}
\end{table}

\begin{figure}[ht]
\centering
\vspace{-0.6\baselineskip}
\begin{center}
\includegraphics[width=0.8\linewidth]{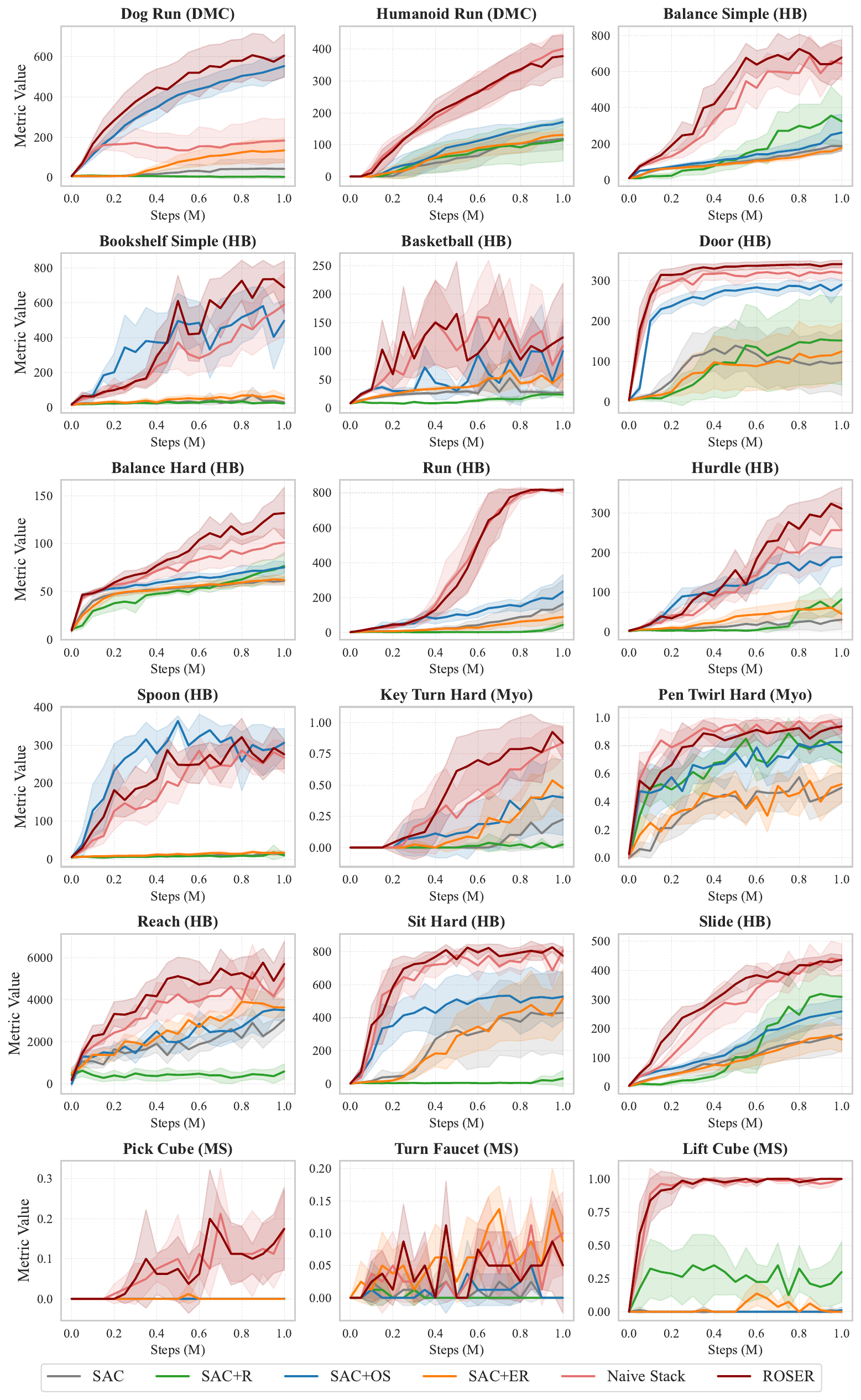}
\end{center}
\caption{
  \textbf{Learning curves across all tasks.} Comparison of \textbf{\texttt{ROSER}} against baseline methods across DMC, HB, Myo, and MS benchmarks.
}
\label{learning_curve_appendix}
\end{figure}

%% file: ref.bib
@inproceedings{haarnoja2018soft,
  title={Soft actor-critic: Off-policy maximum entropy deep reinforcement learning with a stochastic actor},
  author={Haarnoja, Tuomas and Zhou, Aurick and Abbeel, Pieter and Levine, Sergey},
  booktitle={International conference on machine learning},
  pages={1861--1870},
  year={2018},
  organization={Pmlr}
}

@article{agarwal2021deep,
  title={Deep Reinforcement Learning at the Edge of the Statistical Precipice},
  author={Agarwal, Rishabh and Schwarzer, Max and Castro, Pablo Samuel
          and Courville, Aaron and Bellemare, Marc G},
  journal={Advances in Neural Information Processing Systems},
  year={2021}
}

@inproceedings{
    lee2025simba,
    title={SimBa: Simplicity Bias for Scaling Up Parameters in Deep Reinforcement Learning},
    author={Hojoon Lee and Dongyoon Hwang and Donghu Kim and Hyunseung Kim and Jun Jet Tai and Kaushik Subramanian and Peter R. Wurman and Jaegul Choo and Peter Stone and Takuma Seno},
    booktitle={The Thirteenth International Conference on Learning Representations},
    year={2025},
    url={https://openreview.net/forum?id=jXLiDKsuDo}
}

@inproceedings{
lee2025hyperspherical,
title={Hyperspherical Normalization for Scalable Deep Reinforcement Learning},
author={Hojoon Lee and Youngdo Lee and Takuma Seno and Donghu Kim and Peter Stone and Jaegul Choo},
booktitle={Forty-second International Conference on Machine Learning},
year={2025},
url={https://openreview.net/forum?id=kfYxyvCYQ4}
}

@inproceedings{
fujimoto2025towards,
title={Towards General-Purpose Model-Free Reinforcement Learning},
author={Scott Fujimoto and Pierluca D'Oro and Amy Zhang and Yuandong Tian and Michael Rabbat},
booktitle={The Thirteenth International Conference on Learning Representations},
year={2025},
url={https://openreview.net/forum?id=R1hIXdST22}
}

@inproceedings{
fujimoto2023for,
title={For {SALE}: State-Action Representation Learning for Deep Reinforcement Learning},
author={Scott Fujimoto and Wei-Di Chang and Edward J. Smith and Shixiang Shane Gu and Doina Precup and David Meger},
booktitle={Thirty-seventh Conference on Neural Information Processing Systems},
year={2023},
url={https://openreview.net/forum?id=xZvGrzRq17}
}

@misc{bengio2014representation,
      title={Representation Learning: A Review and New Perspectives}, 
      author={Yoshua Bengio and Aaron Courville and Pascal Vincent},
      year={2014},
      eprint={1206.5538},
      archivePrefix={arXiv},
      primaryClass={cs.LG},
      url={https://arxiv.org/abs/1206.5538}, 
}

@article{ha2018world,
  doi = {10.5281/ZENODO.1207631},
  url = {https://zenodo.org/record/1207631},
  author = {Ha, David and Schmidhuber, Jürgen},
  title = {World Models},
  publisher = {Zenodo},
  year = {2018},
  copyright = {Creative Commons Attribution 4.0}
}

@misc{srinivas2020curl,
      title={CURL: Contrastive Unsupervised Representations for Reinforcement Learning}, 
      author={Aravind Srinivas and Michael Laskin and Pieter Abbeel},
      year={2020},
      eprint={2004.04136},
      archivePrefix={arXiv},
      primaryClass={cs.LG},
      url={https://arxiv.org/abs/2004.04136}, 
}

@inproceedings{
Hafner2020Dream,
title={Dream to Control: Learning Behaviors by Latent Imagination},
author={Danijar Hafner and Timothy Lillicrap and Jimmy Ba and Mohammad Norouzi},
booktitle={International Conference on Learning Representations},
year={2020},
url={https://openreview.net/forum?id=S1lOTC4tDS}
}

@article{tassa2018dmc,
  title={Deepmind control suite},
  author={Tassa, Yuval and Doron, Yotam and Muldal, Alistair and Erez, Tom and Li, Yazhe and Casas, Diego de Las and Budden, David and Abdolmaleki, Abbas and Merel, Josh and Lefrancq, Andrew and others},
  journal={arXiv preprint arXiv:1801.00690},
  year={2018}
}

@article{sferrazza2024humanoidbench,
    title={HumanoidBench: Simulated Humanoid Benchmark for Whole-Body Locomotion and Manipulation},
    author={Carmelo Sferrazza and Dun-Ming Huang and Xingyu Lin and Youngwoon Lee and Pieter Abbeel},
    journal={arXiv Preprint arxiv:2403.10506},
    year={2024}
}

@Misc{MyoSuite2022,
  author =       {Vittorio, Caggiano AND Huawei, Wang AND Guillaume, Durandau AND Massimo, Sartori AND Vikash, Kumar},
  title =        {MyoSuite -- A contact-rich simulation suite for musculoskeletal motor control},
  publisher = {arXiv},
  year = {2022},
  howpublished = {\url{https://github.com/myohub/myosuite}},
  doi = {10.48550/ARXIV.2205.13600},
  url = {https://arxiv.org/abs/2205.13600},
}

@inproceedings{gu2023maniskill2,
  title={ManiSkill2: A Unified Benchmark for Generalizable Manipulation Skills},
  author={Gu, Jiayuan and Xiang, Fanbo and Li, Xuanlin and Ling, Zhan and Liu, Xiqiaing and Mu, Tongzhou and Tang, Yihe and Tao, Stone and Wei, Xinyue and Yao, Yunchao and Yuan, Xiaodi and Xie, Pengwei and Huang, Zhiao and Chen, Rui and Su, Hao},
  booktitle={International Conference on Learning Representations},
  year={2023}
}

@inproceedings{hansen2024tdmpc2,
  title={TD-MPC2: Scalable, Robust World Models for Continuous Control}, 
  author={Nicklas Hansen and Hao Su and Xiaolong Wang},
  booktitle={International Conference on Learning Representations (ICLR)},
  year={2024}
}

@article{schaul2015prioritized,
  title={Prioritized experience replay},
  author={Schaul, Tom and Quan, John and Antonoglou, Ioannis and Silver, David},
  journal={arXiv preprint arXiv:1511.05952},
  year={2015}
}

@article{fujimoto2020equivalence,
  title={An Equivalence between Loss Functions and Non-Uniform Sampling in Experience Replay},
  author={Fujimoto, Scott and Meger, David and Precup, Doina},
  journal={Advances in Neural Information Processing Systems},
  volume={33},
  year={2020}
}

@misc{lahire2021large,
      title={Large Batch Experience Replay}, 
      author={Thibault Lahire and Matthieu Geist and Emmanuel Rachelson},
      year={2021},
      eprint={2110.01528},
      archivePrefix={arXiv},
      primaryClass={cs.LG}
}

@article{sujit2022prioritizing,
  title   = {Prioritizing Samples in Reinforcement Learning with Reducible Loss},
  author  = {Shivakanth Sujit and Somjit Nath and Pedro H. M. Braga and Samira Ebrahimi Kahou},
  year    = {2022},
  journal = {arXiv preprint arXiv: Arxiv-2208.10483}
}

@inproceedings{
nauman2024bigger,
title={Bigger, Regularized, Optimistic: scaling for compute and sample-efficient continuous control},
author={Michal Nauman and Mateusz Ostaszewski and Krzysztof Jankowski and Piotr Miłoś and Marek Cygan},
booktitle={Advances in Neural Information Processing Systems},
year={2024},
url={https://arxiv.org/pdf/2405.16158},
}

@misc{klein2024plasticitylossdeepreinforcement,
      title={Plasticity Loss in Deep Reinforcement Learning: A Survey}, 
      author={Timo Klein and Lukas Miklautz and Kevin Sidak and Claudia Plant and Sebastian Tschiatschek},
      year={2024},
      eprint={2411.04832},
      archivePrefix={arXiv},
      primaryClass={cs.AI},
      url={https://arxiv.org/abs/2411.04832}, 
}

@inproceedings{nikishin2022primacy,
  title={The Primacy Bias in Deep Reinforcement Learning},
  author={Nikishin, Evgenii and Schwarzer, Max and D'Oro, Pierluca and Bacon, Pierre-Luc and Courville, Aaron},
  booktitle={International Conference on Machine Learning},
  year={2022},
  organization={PMLR}
}

@inproceedings{
mahankali2024random,
title={Random Latent Exploration for Deep Reinforcement Learning},
author={Srinath V. Mahankali and Zhang-Wei Hong and Ayush Sekhari and Alexander Rakhlin and Pulkit Agrawal},
booktitle={Forty-first International Conference on Machine Learning},
year={2024},
url={https://openreview.net/forum?id=Y9qzwNlKVU}
}

@inproceedings{
sukhija2025maxinforl,
title={MaxInfo{RL}: Boosting exploration in reinforcement learning through information gain maximization},
author={Bhavya Sukhija and Stelian Coros and Andreas Krause and Pieter Abbeel and Carmelo Sferrazza},
booktitle={The Thirteenth International Conference on Learning Representations},
year={2025},
url={https://openreview.net/forum?id=R4q3cY3kQf}
}

@inproceedings{yu2018towards,
  title={Towards Sample Efficient Reinforcement Learning.},
  author={Yu, Yang},
  booktitle={IJCAI},
  pages={5739--5743},
  year={2018}
}

@misc{ma2025rethinkingroledynamicsparse,
      title={Rethinking the Role of Dynamic Sparse Training for Scalable Deep Reinforcement Learning}, 
      author={Guozheng Ma and Lu Li and Zilin Wang and Haoyu Wang and Shengchao Hu and Leszek Rutkowski and Dacheng Tao},
      year={2025},
      eprint={2510.12096},
      archivePrefix={arXiv},
      primaryClass={cs.LG},
      url={https://arxiv.org/abs/2510.12096}, 
}

@misc{fedus2020revisitingfundamentalsexperiencereplay,
      title={Revisiting Fundamentals of Experience Replay}, 
      author={William Fedus and Prajit Ramachandran and Rishabh Agarwal and Yoshua Bengio and Hugo Larochelle and Mark Rowland and Will Dabney},
      year={2020},
      eprint={2007.06700},
      archivePrefix={arXiv},
      primaryClass={cs.LG},
      url={https://arxiv.org/abs/2007.06700}, 
}

@inproceedings{Ansel_PyTorch_2_Faster_2024,
author = {Ansel, Jason and Yang, Edward and He, Horace and Gimelshein, Natalia and Jain, Animesh and Voznesensky, Michael and Bao, Bin and Bell, Peter and Berard, David and Burovski, Evgeni and Chauhan, Geeta and Chourdia, Anjali and Constable, Will and Desmaison, Alban and DeVito, Zachary and Ellison, Elias and Feng, Will and Gong, Jiong and Gschwind, Michael and Hirsh, Brian and Huang, Sherlock and Kalambarkar, Kshiteej and Kirsch, Laurent and Lazos, Michael and Lezcano, Mario and Liang, Yanbo and Liang, Jason and Lu, Yinghai and Luk, CK and Maher, Bert and Pan, Yunjie and Puhrsch, Christian and Reso, Matthias and Saroufim, Mark and Siraichi, Marcos Yukio and Suk, Helen and Suo, Michael and Tillet, Phil and Wang, Eikan and Wang, Xiaodong and Wen, William and Zhang, Shunting and Zhao, Xu and Zhou, Keren and Zou, Richard and Mathews, Ajit and Chanan, Gregory and Wu, Peng and Chintala, Soumith},
booktitle = {29th ACM International Conference on Architectural Support for Programming Languages and Operating Systems, Volume 2 (ASPLOS '24)},
doi = {10.1145/3620665.3640366},
month = apr,
publisher = {ACM},
title = {{PyTorch 2: Faster Machine Learning Through Dynamic Python Bytecode Transformation and Graph Compilation}},
url = {https://docs.pytorch.org/assets/pytorch2-2.pdf},
year = {2024}
}

@inproceedings{hessel2018rainbow,
  title={Rainbow: Combining improvements in deep reinforcement learning},
  author={Hessel, Matteo and Modayil, Joseph and Van Hasselt, Hado and Schaul, Tom and Ostrovski, Georg and Dabney, Will and Horgan, Dan and Piot, Bilal and Azar, Mohammad and Silver, David},
  booktitle={Proceedings of the AAAI conference on artificial intelligence},
  volume={32},
  number={1},
  year={2018}
}

@inproceedings{ceron2021revisiting,
  title={Revisiting rainbow: Promoting more insightful and inclusive deep reinforcement learning research},
  author={Ceron, Johan Samir Obando and Castro, Pablo Samuel},
  booktitle={International Conference on Machine Learning},
  pages={1373--1383},
  year={2021},
  organization={PMLR}
}

@inproceedings{
clark2025beyond,
title={Beyond The Rainbow: High Performance Deep Reinforcement Learning on a Desktop {PC}},
author={Tyler Clark and Mark Towers and Christine Evers and Jonathon Hare},
booktitle={Forty-second International Conference on Machine Learning},
year={2025},
url={https://openreview.net/forum?id=V3KXsUFw8D}
}

@inproceedings{
kong2025mastering,
title={Mastering Massive Multi-Task Reinforcement Learning via Mixture-of-Expert Decision Transformer},
author={Yilun Kong and Guozheng Ma and Qi Zhao and Haoyu Wang and Li Shen and Xueqian Wang and Dacheng Tao},
booktitle={ICLR 2025 Workshop on Modularity for Collaborative, Decentralized, and Continual Deep Learning},
year={2025},
url={https://openreview.net/forum?id=YgR8U5DSj9}
}

@article{
kong2025qpo,
title={{QPO}: Query-dependent Prompt Optimization via Multi-Loop Offline Reinforcement Learning},
author={Yilun Kong and Hangyu Mao and Zhao Qi and Bin Zhang and Jingqing Ruan and Li Shen and Yongzhe Chang and Xueqian Wang and Rui Zhao and Dacheng Tao},
journal={Transactions on Machine Learning Research},
issn={2835-8856},
year={2025},
url={https://openreview.net/forum?id=bqMJToTkvT},
note={}
}
